\documentclass{article} %
\usepackage{iclr2027_conference,times}

\usepackage{amsmath,amsfonts,bm}

\def\1{\bm{1}}

\def\vmu{{\bm{\mu}}}

\def\vb{{\bm{b}}}
\def\vc{{\bm{c}}}
\def\vd{{\bm{d}}}

\def\vf{{\bm{f}}}
\def\vg{{\bm{g}}}

\def\vo{{\bm{o}}}
\def\vp{{\bm{p}}}
\def\vq{{\bm{q}}}

\def\vs{{\bm{s}}}

\def\vu{{\bm{u}}}
\def\vv{{\bm{v}}}

\def\vz{{\bm{z}}}

\def\mA{{\bm{A}}}
\def\mB{{\bm{B}}}
\def\mC{{\bm{C}}}

\def\mF{{\bm{F}}}

\def\mH{{\bm{H}}}

\def\mQ{{\bm{Q}}}

\def\mS{{\bm{S}}}

\def\mW{{\bm{W}}}

\def\mZ{{\bm{Z}}}

\DeclareMathAlphabet{\mathsfit}{\encodingdefault}{\sfdefault}{m}{sl}
\SetMathAlphabet{\mathsfit}{bold}{\encodingdefault}{\sfdefault}{bx}{n}
\newcommand{\tens}[1]{\bm{\mathsfit{#1}}}

\def\tU{{\tens{U}}}
\def\tV{{\tens{V}}}

\def\sR{{\mathbb{R}}}

\usepackage{hyperref}
\usepackage{url}
\usepackage{booktabs}
\usepackage{multirow}
\usepackage{graphicx}
\usepackage{duckuments}
\usepackage{wrapfig}
\usepackage{needspace}
\usepackage{placeins}
\usepackage{amsmath}
\usepackage{amssymb}
\usepackage{xcolor}
\definecolor{skyblue}{RGB}{0,140,220}
\hypersetup{colorlinks=true, citecolor=skyblue, urlcolor=skyblue, linkcolor=red}

\usepackage{xspace}
\usepackage{colortbl}
\usepackage{subcaption}
\usepackage{wrapfig}
\usepackage{tikz}
\usetikzlibrary{positioning,arrows.meta,fit,calc,backgrounds}
\usepackage{fontawesome5}   %
\definecolor{defaultrow}{gray}{0.90}
\definecolor{scenetraj}{HTML}{A600FF}
\definecolor{handtraj}{HTML}{00B000}
\definecolor{frozenice}{HTML}{3DA5E0}   %
\definecolor{sharedspace}{HTML}{A7A5C6} %
\definecolor{querytoken}{HTML}{DB6C79} %
\definecolor{texttoken}{HTML}{E9C46A} %
\definecolor{visiontoken}{HTML}{5EB1BF} %
\definecolor{actiontoken}{HTML}{ED4D6E} %
\definecolor{graphite}{HTML}{3C4046}   %

\newcommand{\dfltrow}[2]{\cellcolor{defaultrow}#1 & \cellcolor{defaultrow}#2 \\}
\newcommand{\dfltrowiii}[3]{\cellcolor{defaultrow}#1 & \cellcolor{defaultrow}#2 & \cellcolor{defaultrow}#3 \\}
\newcolumntype{O}{>{\columncolor{defaultrow}}r}
\newcommand{\solidrules}{%
  \setlength{\aboverulesep}{0pt}%
  \setlength{\belowrulesep}{0pt}%
  \renewcommand{\arraystretch}{1.22}}
\NewDocumentCommand{\abcell}{O{lc} O{2} m m m m}{%
  \footnotesize\setlength{\tabcolsep}{2.5pt}\renewcommand{\arraystretch}{1.15}%
  \begin{tabular}[t]{#1}
  \multicolumn{#2}{c}{(#3) #4} \\[2pt]
  \toprule
  #5 \\
  \midrule
  #6
  \bottomrule
  \end{tabular}}

\def\cP{{\mathcal{P}}}
\def\cQ{{\mathcal{Q}}}

\newcommand{\methodname}{PointWAM\xspace}
\newcommand{\ie}{\textit{i.e.}\xspace}

\def\vsigma{{\bm{\sigma}}}


\newcommand{\sd}[1]{\raisebox{-0.3ex}{\tiny #1}}

\title{\methodname: 3D World Action Modeling\\
for Dexterous Robotic Manipulation}

\newif\ifarxiv
\arxivtrue
\author{Chunghyun Park$^{1}$\thanks{Co-first authors.\quad$^{\dagger}$Equal contribution.}\quad\hspace{4.58pt}Beomjun Kim$^{2*}$\quad Seungcheol Park$^{4\dagger}$\quad Heeseung Kwon$^{4\dagger}$\\
\textbf{Yashu Shukla}$^{3,4}$\quad \textbf{Seunghoon Sim}$^{2}$\quad \textbf{Jinwoo Shin}$^{2,4}$\quad \textbf{Minsu Cho}$^{1,4}$\\
{\normalfont $^{1}$POSTECH\quad $^{2}$KAIST\quad $^{3}$Hanyang University\quad $^{4}$RLWRLD}}

\ifarxiv\iclrfinalcopy\fi %

\begin{document}

\maketitle
\ifarxiv\lhead{Preprint}\fi

\begin{abstract}
World action models jointly learn to forecast world dynamics and predict robot actions, such that the learned internal world dynamics guide accurate actions.
Existing approaches typically represent the world as RGB frames or latent counterparts while predicting actions as end-effector poses or joint angles, but they often struggle to capture the 3D spatial structure and contact geometry central to dexterous manipulation.
We introduce Point World Action Model (\methodname), a 3D world action model that decomposes the world into a scene (\ie, environment) and hands (\ie, actor), and jointly forecasts both as 3D point trajectories within a shared space-time coordinate frame.
This explicit, disentangled representation enables effective pre-training on large-scale human demonstration videos without requiring any task-specific object or keypoint selection.
Given a colored point cloud and a language instruction, \methodname predicts how the scene and hands co-evolve in 3D space over time, then retargets the forecast hand motion to robot actions.
Pre-training on human videos improves average DexJoCo success by 56.9 percentage points, and scene-trajectory supervision adds 10.9 points over forecasting the hands alone.
With both, \methodname surpasses the prior state of the art on ten DexJoCo tasks by 11.7 points and outperforms strong VLAs on a real robot.
\ifarxiv\par\smallskip\noindent Project page: \url{https://chrockey.github.io/PointWAM}\fi %

\end{abstract}

\begin{figure}[!ht]
\centering
\colorlet{band}{graphite!22}\colorlet{accent}{graphite}
\begin{tikzpicture}[
  font=\fontfamily{phv}\selectfont\scriptsize,
  ph/.style={draw=black!45, dashed, rounded corners=3pt, fill=white, text=black!50,
             font=\fontfamily{phv}\selectfont\scriptsize\itshape, align=center, inner sep=0pt},
  hdr/.style={font=\fontfamily{phv}\selectfont\scriptsize, text=black, inner sep=0pt,
              anchor=base, align=center},
]
\def\tw{1.244}\def\th{0.6998}\def\dx{1.304}
\fill[band, rounded corners=3pt] (0,1.9756) rectangle (6.62,5.1948);
\fill[band, rounded corners=3pt] (0,0) rectangle (6.62,1.6196);
\foreach \i in {0,...,4}{
  \node[anchor=south west, inner sep=0pt] at ({0.08+\i*\dx},4.4150) {\includegraphics[width=\tw cm,height=\th cm]{figures/teaser/rgb\i.jpg}};
  \node[anchor=south west, inner sep=0pt] at ({0.08+\i*\dx},3.6552) {\includegraphics[width=\tw cm,height=\th cm]{figures/teaser/depth\i.jpg}};}
\foreach \i in {5,...,9}{
  \node[anchor=south west, inner sep=0pt] at ({0.08+(\i-5)*\dx},2.8154) {\includegraphics[width=\tw cm,height=\th cm]{figures/teaser/rgb\i.jpg}};
  \node[anchor=south west, inner sep=0pt] at ({0.08+(\i-5)*\dx},2.0556) {\includegraphics[width=\tw cm,height=\th cm]{figures/teaser/depth\i.jpg}};}
\foreach \i in {10,...,14}{
  \node[anchor=south west, inner sep=0pt] at ({0.08+(\i-10)*\dx},0.8398) {\includegraphics[width=\tw cm,height=\th cm]{figures/teaser/rgb\i.jpg}};
  \node[anchor=south west, inner sep=0pt] at ({0.08+(\i-10)*\dx},0.08) {\includegraphics[width=\tw cm,height=\th cm]{figures/teaser/depth\i.jpg}};}
\node[hdr] at (3.31,5.2828) {\textbf{Large-scale pre-training} on 1.15M human demo episodes};
\node[hdr] at (3.31,1.7156) {\textbf{Transfer learning} to dexterous robotic manipulation};

\fill[band] (6.71,2.3774) -- (6.85,2.3774) -- (6.85,2.1974) -- (7.03,2.5974) -- (6.85,2.9974)
  -- (6.85,2.8174) -- (6.71,2.8174) -- cycle;

\fill[band, rounded corners=3pt] (7.06,0) rectangle (13.97,5.1948);
\node[hdr] at (10.515,5.2828) {\textbf{{\color{scenetraj}Scene} and {\color{handtraj}hands}} forecast as 3D point trajectories};
\node[anchor=south, inner sep=0pt] (fc) at (10.515,0.08)
  {\includegraphics[width=6.75cm,height=5.0348cm]{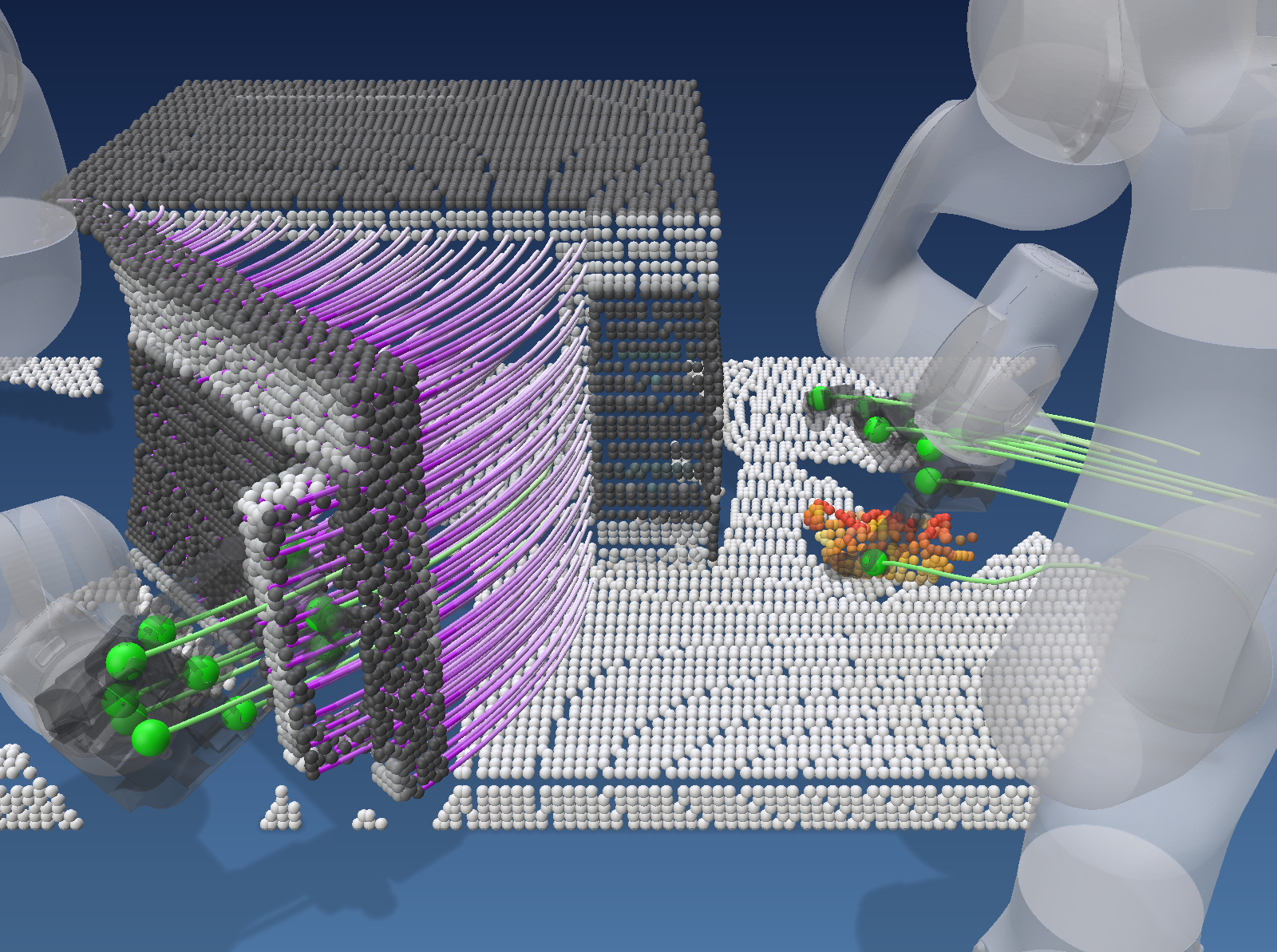}};
\end{tikzpicture}
\caption{\textbf{\methodname forecasts the scene and hands as 3D point trajectories.}
\methodname pre-trains on human demonstration videos and fine-tunes on robot demonstrations, both
expressed as 3D point trajectories (left; depth below RGB). Given a language instruction, its {\color{scenetraj}scene}
and {\color{handtraj}hand} forecasts share one space-time coordinate frame (right), and it
retargets the hand trajectories to robot actions.}
\label{fig:teaser}
\end{figure}

\section{Introduction}
\label{sec:intro}

Dexterous robotic manipulation requires multiple fingers to precisely coordinate their 3D motion relative to object surfaces~\citep{salisbury1982articulated,okamura2000dexterous}, enabling robots to perform a wide range of tasks from grasping objects of varied shapes to tool use~\citep{bicchi2000hands,dexilsurvey2025}. Conventional Vision-Language-Action models (VLAs)~\citep{pi05_2025,groot_n1_2025} map observations to motor commands without explicitly forecasting how the scene evolves. World Action Models (WAMs) have recently emerged as a promising alternative by treating world and action modeling as deeply intertwined: they jointly learn to forecast how the world will change and what actions cause those changes~\citep{uva2025}.

Many existing WAMs build on video generation backbones and represent the world as RGB frames~\citep{unipi2023} or their latent counterparts~\citep{uva2025,genieenvisioner2025}. These representations are limited in their ability to capture key aspects of physical interaction, including 3D spatial structure and causal relations between the actor and the environment. Point-based policies~\citep{pointpolicy2025,dexpointpolicy2026} address the 3D action prediction problem by forecasting trajectories of 3D points, but do not model how the environment evolves. They also rely on humans or external perception models to select task-relevant objects for their scene inputs. Conversely, action-conditioned 3D world models forecast dense 3D scene trajectories, yet treat robot actions as inputs rather than quantities to be predicted~\citep{pointworld2026}.

In this work, we introduce \emph{Point World Action Model} (\methodname), a 3D world action model that decomposes the world into a scene (\ie, environment) and hands (\ie, actor), and represents both as three-dimensional point trajectories within a shared space-time coordinate frame (Fig.~\ref{fig:teaser}). This explicit, disentangled world representation enables effective learning of cause-and-effect relations between the environment and the actor by modeling how their points co-evolve in 3D space over time. Given a colored point cloud and a language instruction, \methodname forecasts trajectories for every observed scene point and hand keypoint in a common coordinate frame. Its action retargeter then converts the forecast hand trajectories into the corresponding robot actions.

The proposed approach offers three additional advantages. First, modeling both the scene and hands in a shared 3D space enables effective pre-training on human demonstration videos from EgoDex~\citep{egodex2026} and VITRA~\citep{vitra2025}. Second, no task-specific objects or scene keypoints need to be selected in advance, so scene forecasting is not restricted to a predefined set of target objects. Third, by decoupling hand-point trajectories from the scene, the forecast hand motion is amenable to effective retargeting onto robot hands for action prediction.

\needspace{9\baselineskip}
Our main contributions can be summarized as follows.
\begingroup
\setlength{\leftmargini}{1.2em}
\begin{itemize}
  \item We introduce \methodname, which jointly forecasts the 3D trajectories of scene points and hand keypoints in a shared space-time coordinate frame, unifying world and action modeling.
  \item We demonstrate that 3D world modeling benefits dexterous manipulation. Scene-trajectory supervision improves average DexJoCo success by 10.9 percentage points over the same model without it, and our 3D encoder outperforms lifted image features by 5.4 percentage points.
  \item We show that pre-training on human demonstrations from EgoDex and VITRA improves average DexJoCo success by 56.9 percentage points over training from scratch. With this pre-training, \methodname surpasses the prior best multi-task success rate on DexJoCo by 11.7 percentage points and outperforms robot-pre-trained VLAs on a real robot by up to 25 percentage points.
\end{itemize}
\endgroup

\section{Related Work}
\label{sec:related}

We discuss the closest work here and provide a broader review in Appendix~\ref{app:related}.

\paragraph{World action models (WAMs).}
WAMs unify predictive world modeling with action prediction; unlike conventional VLAs, which map a visual observation straight to a motor command, a WAM is trained to forecast how the world will change and predict what actions the robot must perform
~\citep{unipi2023,vpp2025,genieenvisioner2025,cosmospolicy2026,uva2025}. Existing WAMs
typically forecast future visual representations in the latent space of pretrained video models
while predicting robot actions as end-effector poses.
These models must therefore learn how changes in latent visual features correspond to robot
motion, rather than explicitly modeling both scene and hand motion as 3D trajectories. Closest to ours,
EgoWAM~\citep{egowam2026}, PointWorld~\citep{pointworld2026}, and PointZero~\citep{pointzero2026}
forecast dense 3D scene motion.
EgoWAM jointly learns scene forecasting and action prediction but maps human hand motion to the
actions of parallel-jaw grippers, a 6-DoF pose and a gripper state per hand, which lacks
individual finger motions. Pri4R~\citep{pri4r2026} likewise
targets grippers, adding a training-only 3D point-trajectory head to a VLA fine-tuned on robot
demonstrations. PointWorld forecasts scene motion with a
3D point-cloud network~\citep{wu2024ptv3} conditioned on future robot trajectories, so it needs a
separate planner that selects them with a task cost. PointZero pre-trains a similar
world model without robot data and, like PointWorld, predicts no actions. \methodname instead pre-trains on human demonstration videos for dexterous
hands, jointly forecasts scene and hand trajectories in a shared space-time coordinate frame, and
retargets the hand trajectories to robot actions without a separate planner.

\paragraph{Point trajectories from human demonstration videos.}
Human demonstration videos show hand and scene motion from which trajectory supervision can be
extracted without recorded robot actions. Point-based policies can learn to forecast hand keypoint
trajectories from these videos and retarget the forecasts to robot actions. Point
Policy~\citep{pointpolicy2025} and Dexterous Point Policy~\citep{dexpointpolicy2026}, however, rely
on humans or external perception models to select the objects in the policy's scene input, object
keypoints specified by a human for Point Policy and object point clouds segmented by external models
for Dexterous Point Policy. Neither performs explicit world modeling through scene-motion
forecasting. 3PoinTr~\citep{threepointr2026} does forecast dense 3D scene trajectories from human
demonstration videos. Extracted
video tracks supervise its forecasts, and a separate policy trained on robot demonstrations predicts
actions from the forecast tracks. \methodname instead forecasts trajectories for every observed scene
point and hand keypoint with one backbone, and a retargeter converts the hand trajectories into
robot actions without a separate policy.

\section{Point World Action Model}
\label{sec:method}

\begin{figure}[!t]
\centering
\begin{tikzpicture}[
  font=\fontfamily{phv}\selectfont\scriptsize, >={Stealth[length=3pt]},
  every node/.style={align=center, inner sep=1.5pt},
  blk/.style={draw=black!70, rounded corners=1.5pt, fill=black!7, line width=0.45pt,
              font=\fontfamily{phv}\selectfont\scriptsize},
  frz/.style={blk},  %
  viz/.style={draw=black!40, dashed, rounded corners=2pt, fill=black!3, text=black!50,
              font=\fontfamily{phv}\selectfont\tiny\itshape},
  lab/.style={font=\fontfamily{phv}\selectfont\scriptsize},
  ttl/.style={font=\fontfamily{phv}\selectfont\small\bfseries, anchor=south west, inner sep=0pt},
  sub/.style={font=\fontfamily{phv}\selectfont\scriptsize, text=black!55, anchor=base west, inner sep=0pt},
  frozen/.style={font=\tiny, text=frozenice, inner sep=0pt},
  arr/.style={->, line width=0.5pt},
  at box/.style args={#1,#2,#3,#4}{at={(#1,#2)}, minimum width=#3cm, minimum height=#4cm},
]
\fill[sharedspace!30, rounded corners=4pt] (2.75,1.25) -- (8.425,1.25) -- (8.425,0.35) -- (10.75,0.35)
  -- (10.75,4.75) -- (2.75,4.75) -- cycle;
\node[sub, anchor=base east, text=sharedspace!55!black] at (10.75,4.88) {\itshape shared space-time coordinate frame};
\node[at box={0.85,3.8,1.7,1.6}, inner sep=0pt] (pc)
  {\includegraphics[width=1.7cm,height=1.6cm]{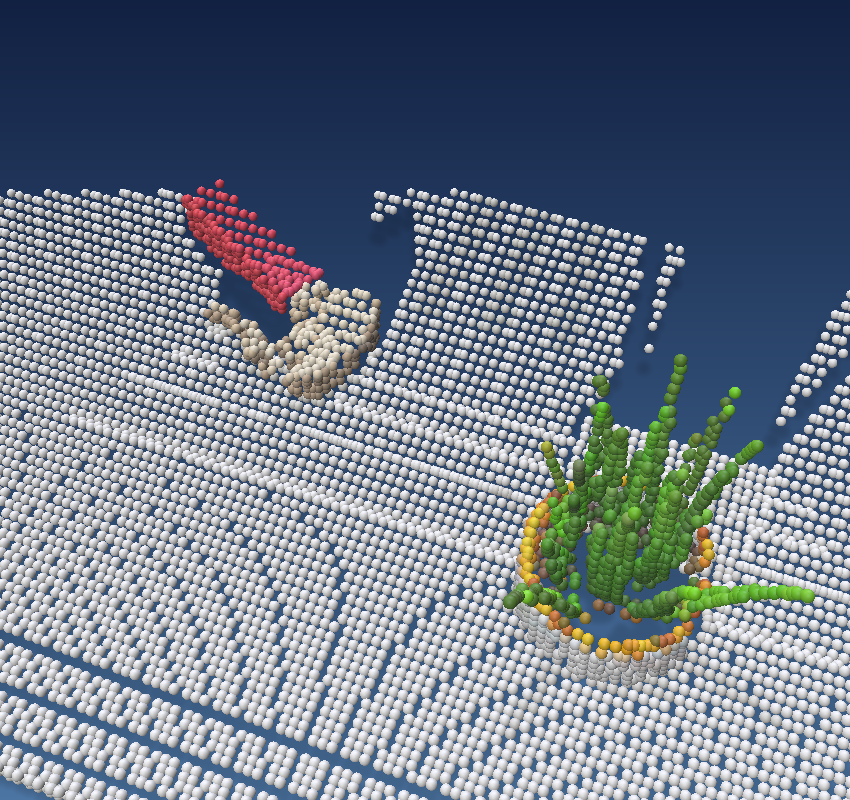}};
\node[lab, anchor=north, inner sep=0pt] at ([yshift=-0.1cm]pc.south) {scene points};
\node[frz, at box={2.3,3.8,0.5,1.6}]    (enc) {\rotatebox{90}{3D Encoder}};
\node[frozen] at (enc.north west) {\faSnowflake};  %

\node[at box={0.85,2.075,1.7,0.75}, inner sep=0pt] (kp)
  {\includegraphics[width=1.7cm,height=0.75cm]{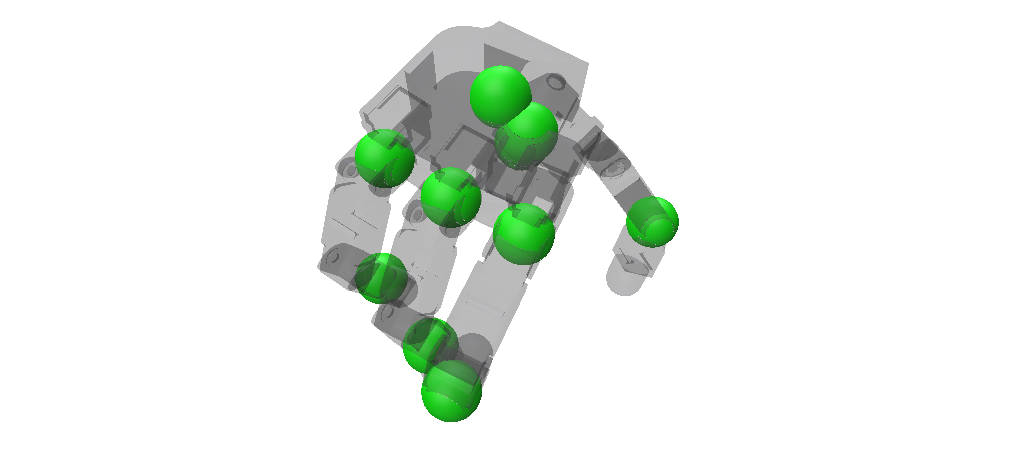}};
\node[lab, anchor=north, inner sep=0pt] at ([yshift=-0.1cm]kp.south) {hand keypoints};
\node[blk, at box={2.3,2.075,0.5,0.75}] (mlp) {\rotatebox{90}{MLP}};

\node[anchor=west, inner xsep=0pt, minimum height=0.6cm, align=left,
      font=\fontfamily{phv}\selectfont\tiny\itshape] (ins) at (0,0.8)
  {``Grasp the watering can and\\apply water to the plant.''};
\path let \p1=(ins.east) in coordinate (tx0) at (\x1+0.45cm,0.8);
\path let \p1=(tx0) in node[frz, minimum height=0.55cm, minimum width={5.39cm-\x1}, anchor=west] (txt) at (tx0) {Text Encoder};
\node[frozen] at (txt.north east) {\faSnowflake};

\node[blk, at box={3.15,3.15,0.5,2.9}]  (tok) {\rotatebox{90}{3D Tokenization}};
\node[at box={5.525,3.225,3.55,2.75}, inner sep=0pt] (vt) {};
\node[at box={5.125,3.15,2.75,2.9}, inner sep=0pt] (vz)
  {\includegraphics[width=2.75cm,height=2.9cm]{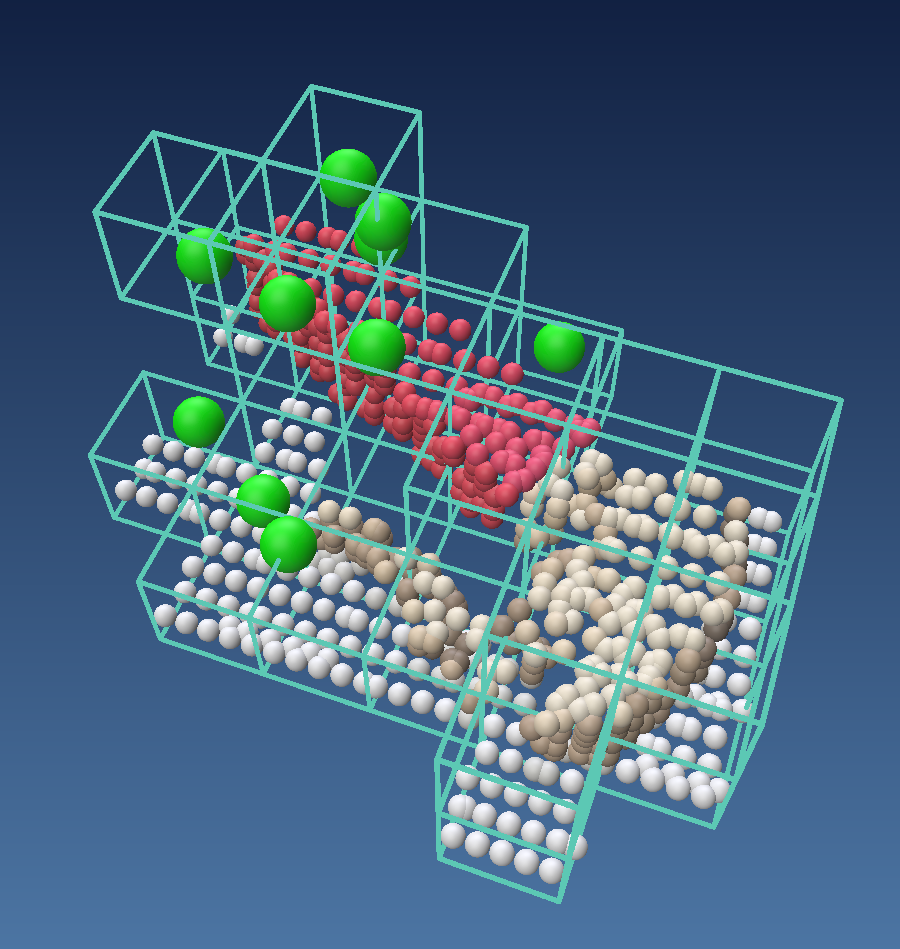}};
\draw[black, line width=0.45pt] (vz.south west) rectangle (vz.north east);
\tikzset{vtok/.style={draw=visiontoken!70!black, fill=visiontoken!60, minimum size=0.26cm, inner sep=0pt},
         vtokh/.style={vtok}}  %
\node[vtokh] (vt1) at (7.075,4.47) {};
\node[vtokh] (vt2) at (7.075,4.11) {};
\foreach \y in {3.86,3.76,3.66}{\fill[black!70] (7.075,\y) circle (0.022);}
\node[vtokh] (vt3) at (7.075,3.39) {};
\node[vtok] (vt4) at (7.075,3.03) {};
\draw[black, line width=0.45pt, line cap=round] (vt1.north west) -- (vz.north east) (vt3.south west) -- (vz.south east);
\node[lab, anchor=north, inner sep=0pt] (vzl) at ([yshift=-0.1cm]vz.south) {zoomed-in view};
\node[lab, anchor=south, inner sep=0pt, align=center] (vtl) at (7.075,0 |- vzl.south) {vision\\tokens};
\node[vtok, anchor=south] (vtL) at ([yshift=0.1cm]vtl.north) {};
\foreach \f in {0.3,0.5,0.7}{\fill[black!70] ($(vt4.south)!\f!(vtL.north)$) circle (0.022);}
\foreach \x/\n in {5.93/tt1,6.29/tt2,7.17/ttL}{\node[draw=texttoken!70!black, fill=texttoken!60, minimum size=0.26cm, inner sep=0pt] (\n) at (\x,0.8) {};}
\foreach \x in {6.63,6.73,6.83}{\fill[black!70] (\x,0.8) circle (0.022);}
\node[lab, anchor=south, inner sep=0pt] at (6.55,0.4) {text tokens};
\node[blk, at box={7.95,2.55,0.6,4.1}]  (tf)  {\rotatebox{90}{Transformer Encoder}};
\node[blk, at box={9.6,4.375,2.0,0.45}] (unp) {Detokenization};
\node[blk, at box={9.0875,3.725,0.975,0.45}] (hs) {MLP$_{\color{scenetraj}\vu}$};
\node[blk, at box={10.1125,3.725,0.975,0.45}] (hh) {MLP$_{\color{handtraj}\vv}$};
\node[at box={9.6,1.9,2.0,2.8}, inner sep=0pt] (fc)
  {\includegraphics[width=2.0cm,height=2.8cm]{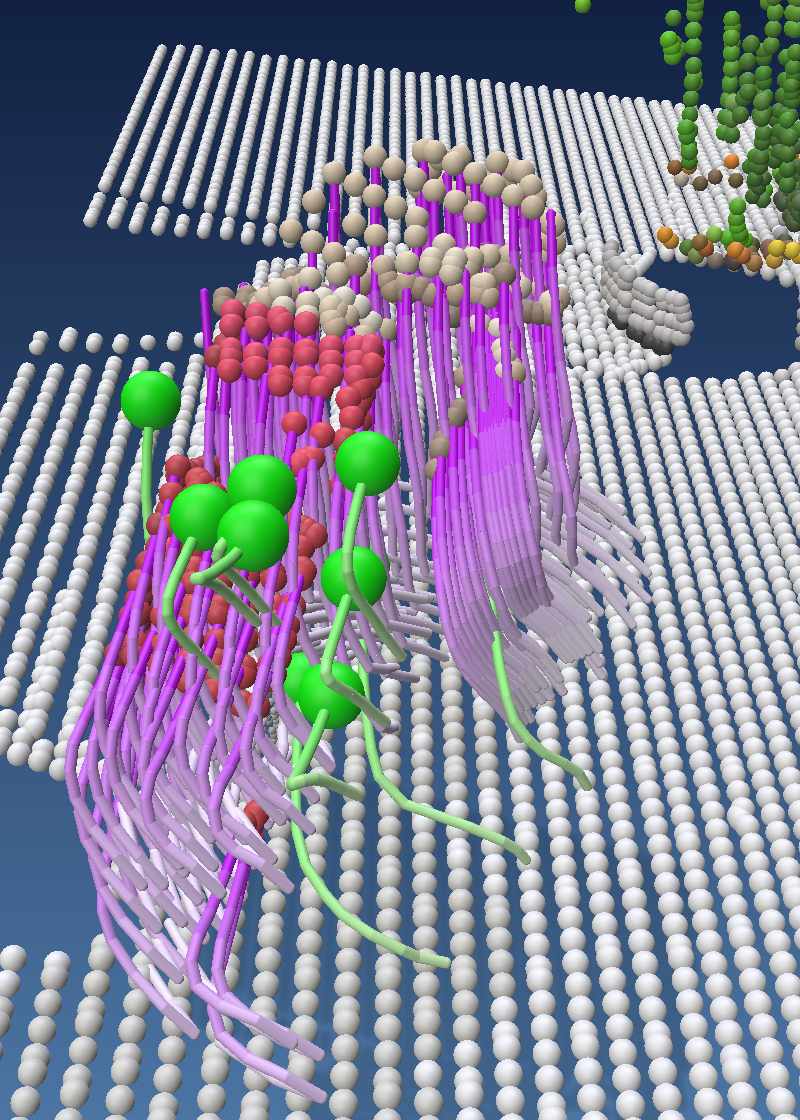}};
\draw[black, line width=0.45pt] (fc.south west) rectangle (fc.north east);  %

\draw[arr] (pc) -- (enc);  \draw[arr] (enc.east) -- (enc.east -| tok.west);
\draw[arr] (kp) -- (mlp);  \draw[arr] (mlp.east) -- (mlp.east -| tok.west);
\draw[arr] ([xshift=0.1cm]ins.east) -- (txt); \draw[arr] (txt.east |- 0,0.8) -- ([xshift=-0.06cm]tt1.west |- 0,0.8);  %
\draw[arr] (tok.east |- 0,3.05) -- (vt.west |- 0,3.05);
\draw[arr] (vt.east |- 0,3.05) -- (tf.west |- 0,3.05);
\draw[arr] ([xshift=0.06cm]ttL.east |- 0,0.8) -- (tf.west |- 0,0.8);
\draw[arr] (tf.east |- unp.west) -- (unp.west);
\draw[arr] (unp.south -| hs) -- (hs.north); \draw[arr] (unp.south -| hh) -- (hh.north);
\draw[arr] (hs.south) -- (hs.south |- fc.north); \draw[arr] (hh.south) -- (hh.south |- fc.north);
\node[ttl, anchor=base west] (t1) at (0,4.88) {3D Trajectory Forecaster};

\draw[black!40, dashed, line width=0.5pt] (10.95,0.45) -- (10.95,5.05);

\tikzset{tok/.style={minimum size=0.26cm, inner sep=0pt},
         ctok/.style={minimum size=0.34cm, inner sep=0pt,
                      font=\fontfamily{phv}\selectfont\tiny}}
\node[draw=handtraj!80, fill=handtraj!15, ctok] (c1) at (11.53,3.57) {$\vv_1$};
\node[draw=handtraj!80, fill=handtraj!15, ctok] (c2) at (11.53,3.11) {$\vv_2$};
\foreach \y in {2.82,2.72,2.62}{\fill[black!70] (11.53,\y) circle (0.022);}
\node[draw=handtraj!80, fill=handtraj!15, ctok] (cT) at (11.53,2.33) {$\vv_T$};
\node[draw=black!60, fill=black!12, ctok] (stk) at (11.53,1.87) {$\vs_t$};
\node[draw=black!45, rounded corners=2pt, fit=(c1)(stk), inner sep=1.5pt] (ctx) {};
\node[lab, anchor=north, inner sep=0pt] at ([yshift=-0.14cm]ctx.south) {context\\$\mC_t$};
\path let \p1=(ctx.east), \p2=(ctx.north), \p3=(ctx.south) in node[blk, anchor=west,
  minimum width={13.86cm-\x1-0.35cm}, minimum height=1.686cm] (dec) at (\x1+0.35cm,{(\y2+\y3)/2})
  {Transformer\\Decoder};
\draw[arr] (ctx.east) -- (dec.west);
\path let \p1=(dec.center), \p2=(dec.south) in coordinate (qc) at (\x1,\y2-0.35cm-0.23cm);  %
\foreach \dx in {-0.72,-0.30,0.72}{\node[draw=querytoken!70!black, fill=querytoken!35, tok] at ($(qc)+(\dx,0)$) {};}
\node[inner sep=0pt] at ($(qc)+(0.21,0)$) {$\cdots$};
\node[lab, anchor=north, inner sep=0pt] at ($(qc)+(0,-0.228)$) {\begin{tabular}{@{}c@{}}learnable\\queries\end{tabular}\,$\mQ$};  %
\draw[arr] ($(qc)+(0,0.23)$) -- (dec.south);  %
\path let \p1=(dec.center), \p2=(dec.north) in coordinate (ac) at (\x1,\y2+0.35cm+0.23cm);  %
\foreach \dx in {-0.72,-0.30,0.72}{\node[draw=actiontoken!70!black, fill=actiontoken!70, tok] at ($(ac)+(\dx,0)$) {};}  %
\node[inner sep=0pt] at ($(ac)+(0.21,0)$) {$\cdots$};
\node[lab, anchor=south, inner sep=0pt] (act) at ($(ac)+(0,0.228)$) {action chunk $\mA_t$};
\draw[arr] (fc.east |- ctx.west) -- (ctx.west);  %
\draw[arr] (dec.north) -- ($(ac)+(0,-0.23)$);  %
\node[ttl, anchor=base west] (t2) at (11.3,4.88) {Action Retargeter};
\end{tikzpicture}
\caption{\textbf{\methodname architecture.} The 3D trajectory forecaster (left) encodes scene
points, hand keypoints, and a language instruction, and jointly forecasts scene
${\color{scenetraj}\vu}$ and hand ${\color{handtraj}\vv}$ trajectories in a shared space-time
coordinate frame. The action retargeter (right) maps ${\color{handtraj}\vv}$ and the robot state
$\vs_t$ to an action chunk $\mA_t$. MLP denotes a multilayer perceptron, and
{\color{frozenice}\faSnowflake} marks frozen modules.}
\label{fig:method}
\end{figure}

We first formulate the task and outline our approach (Section~\ref{sec:formulation}). We then
describe the trajectory forecaster and action retargeter shown in Fig.~\ref{fig:method}
(Sections~\ref{sec:architecture} and~\ref{sec:retargeter}), followed by trajectory supervision and
two-stage training on human and robot demonstrations (Section~\ref{sec:training}).

\subsection{Problem Statement and Overview}
\label{sec:formulation}

Given an instruction $\ell$, a scene point cloud $\cP_t$, and the current robot state $\vs_t$,
the goal is to predict an action chunk $\mA_t\in\sR^{T\times d_a}$ that carries out $\ell$,
where $T$ is the horizon and $d_a$ the action dimension.
The point cloud $\cP_t=\{(\vp^{(i)}_t,\vc^{(i)}_t)\}_{i=1}^{N}$ contains 3D positions and colors
from RGBD frames, while $\vs_t$ records the end-effector pose and finger joint angles of each arm.

We approach this problem by jointly forecasting scene and hand trajectories and retargeting the
hand trajectories to robot actions. To establish a common representation for both human and robot hands,
we use semantically corresponding 3D keypoints, as in point-based policies~\citep{pointpolicy2025,dexpointpolicy2026}.
Specifically, we represent the hands with $K$ keypoints, ten per hand, as
$\cQ_t=\{\vq^{(k)}_t\}_{k=1}^{K}$ in the scene's coordinate frame. Their positions come from robot
states or tracked human hands (Section~\ref{sec:preprocessing}). We express both trajectories as displacements from time $t$:
\begin{equation}
  \vu^{(i)}_{t,\tau} = \vp^{(i)}_{t+\tau} - \vp^{(i)}_t,
  \qquad
  \vv^{(k)}_{t,\tau} = \vq^{(k)}_{t+\tau} - \vq^{(k)}_t,
  \qquad \tau=1,\ldots,T.
  \label{eq:motion_target}
\end{equation}
Scene trajectories supervise world modeling, and hand trajectories serve as an intermediate action
representation.
A trajectory forecaster $f(\cdot)$ predicts both from the scene, hand keypoints, and instruction,
and an action retargeter $g(\cdot)$ maps forecast hand trajectories and robot state to actions:
\begin{equation}
  (\tU_t,\tV_t) = f(\cP_t,\cQ_t,\ell),
  \qquad
  \mA_t = g(\tV_t,\vs_t).
  \label{eq:retarget}
\end{equation}
Here $\tU_t\in\sR^{N\times T\times 3}$ and $\tV_t\in\sR^{K\times T\times 3}$ stack the
forecast scene and hand displacements, which we standardize per horizon step for
training (Appendix~\ref{app:normalization}).

\subsection{3D Trajectory Forecaster}
\label{sec:architecture}

\textbf{Scene and hand features.} We first encode the observed scene point cloud and hand keypoints. A frozen Mosaic3D encoder~\citep{mosaic3d2025}, pre-trained
to align 3D point features with text, extracts a feature vector for each scene point
that captures local geometry and semantics. For each hand keypoint, a multilayer perceptron (MLP) maps a one-hot encoding of its index,
concatenated with a hand ID, to a feature of the same dimension as the scene features.

\textbf{Tokenization.} We place scene and hand features on a shared 1\,cm voxel grid using
their 3D coordinates, then aggregate neighboring features with a sparse-convolutional stem.
We retain these stem features as $\mS_t$ for decoding.
Following Volt~\citep{yilmaz2026volt}, we divide this grid into non-overlapping patches of
$K_v^3$ voxels. For each nonempty patch, we concatenate its voxel features in a fixed
spatial order and linearly project them into one vision token
$\vz^{v}_j\in\sR^d$, $j=1,\ldots,M$. A saved mapping $\mathcal{I}_t$ records the patch and voxel of each point.
We linearly project the $L$ features of the instruction $\ell$ from a frozen
text encoder~\citep{li2024recaption} into instruction tokens
$\vz^{\ell}_1,\ldots,\vz^{\ell}_L$.

\textbf{Multimodal fusion and trajectory forecasting.} We concatenate the instruction and vision
tokens and process them with a transformer encoder backbone using global self-attention~\citep{vaswani2017attention}
and axial 3D rotary position embeddings~\citep{su2024roformer,yilmaz2026volt}, allowing each vision
token to attend to all instruction and vision tokens:
\begin{equation}
  [\mZ^{\ell}_t;\mZ^{v}_t]
  = \mathrm{TransformerEncoder}\big(
    [\vz^{\ell}_1,\ldots,\vz^{\ell}_L,\vz^{v}_1,\ldots,\vz^{v}_M]^\top\big).
  \label{eq:joint_attention}
\end{equation}
We then recover per-point features from the output vision tokens and combine them with the stem
features $\mS_t$ through a residual MLP over their concatenation $[\cdot;\cdot]$, which retains
local detail:
\begin{equation}
  \mF_t = \mF^{\prime}_t
    + \operatorname{MLP}([\mF^{\prime}_t;\mS_t]),
  \qquad \text{where}\quad
  \mF^{\prime}_t = \operatorname{Detokenizer}(\mZ^{v}_t;\mathcal{I}_t).
  \label{eq:trajectory_decoder}
\end{equation}
Here $\operatorname{Detokenizer}$ uses a linear layer to expand each token into $K_v^3$ voxel
features, then selects the feature for each point through $\mathcal{I}_t$.
We then split $\mF_t$ into scene and hand features, $\mF^{\mathrm{scene}}_t$ and $\mF^{\mathrm{hand}}_t$,
and forecast their trajectories with two MLP heads:
\begin{equation}
  \tU_t = \operatorname{MLP}_{\vu}(\mF^{\mathrm{scene}}_t),
  \qquad
  \tV_t = \operatorname{MLP}_{\vv}(\mF^{\mathrm{hand}}_t).
  \label{eq:trajectory_heads}
\end{equation}
The scene head forecasts a trajectory for every point in $\cP_t$, so it requires no task-specific
selection of target objects or keypoints. More importantly, joint supervision encourages the shared backbone to learn
how scene and hand points co-evolve in 3D over time.

\subsection{Action Retargeter}
\label{sec:retargeter}

The action retargeter maps forecast hand trajectories and the robot state to $T$ robot actions;
scene forecasts supervise the shared backbone but do not enter the retargeter.
To construct its input, we encode the hand trajectories at each future step, add a step embedding,
and append a state embedding to form a context $\mC_t$ for action prediction. A transformer
decoder~\citep{vaswani2017attention} then uses this context and learnable action queries to
predict the action chunk $\mA_t$:
\begin{equation}
  \begin{aligned}
  \mC_t &= \big[\operatorname{MLP}_{\mathrm{disp}}(\tV_{t,1})+\vb_1,\;\ldots,\;
            \operatorname{MLP}_{\mathrm{disp}}(\tV_{t,T})+\vb_T,\;\operatorname{MLP}_{\mathrm{state}}(\vs_t)\big]^\top
         \in\sR^{(T+1)\times d_r},\\
  \mA_t &= \mathrm{TransformerDecoder}(\mQ,\mC_t)\in\sR^{T\times d_a}.
  \end{aligned}
  \label{eq:action_retargeter}
\end{equation}
Here $\tV_{t,\tau}\in\sR^{K\times 3}$ contains the hand displacements at future step $\tau$,
and $\vb_\tau$ is a learned step embedding; both MLPs output $d_r$-dimensional features.
Each decoder block applies self-attention among the learnable action queries $\mQ\in\sR^{T\times d_r}$,
cross-attention to $\mC_t$, and a feed-forward network.
A final linear layer maps the features to $T$ actions of end-effector poses and finger joint
targets.

\Needspace{6\baselineskip}
\subsection{Learning from Human and Robot Demonstrations}
\label{sec:preprocessing}
\label{sec:training}

We train the trajectory forecaster on human and robot demonstrations and the action retargeter on
robot demonstrations alone, since only they record the end-effector poses and finger joint commands
that the action retargeter learns to predict (Fig.~\ref{fig:teaser}, left).

\textbf{Human demonstrations.} Human demonstration videos lack recorded states, so we estimate
both trajectories from the videos themselves. Hand trajectories come from the 3D hand poses annotated
in EgoDex~\citep{egodex2026} and reconstructed by VITRA~\citep{vitra2025}, mapped to the robot
hand's ten keypoints. Scene trajectories come from dense 2D point tracks of
CoTracker3~\citep{karaev2025cotracker3}, lifted to 3D with monocular depth from
VGGT-$\Omega$~\citep{wang2026vggtomega} scaled to the annotated hands. The resulting 1.15M episodes share the trajectory format of
the robot demonstrations, so both supervise the same forecaster; Appendix~\ref{app:human_preprocessing} details the preprocessing.

\textbf{Robot demonstrations.} Recorded robot states give hand trajectories through forward
kinematics. Scene trajectories come from replaying DexJoCo~\citep{dexjoco2026} demonstrations,
which move each initial point with its object or rigid part, and on the real robot
from CoTracker3~\citep{karaev2025cotracker3} tracks lifted with Fast-FoundationStereo~\citep{ffs2025} depth. We process the
RoboDojo-Precision demonstrations similarly to DexJoCo (Appendix~\ref{app:supervision}).

\textbf{Training.} We first pre-train the backbone and both trajectory heads from scratch on
human demonstrations using scene and hand trajectory supervision.
On robot demonstrations, we then fine-tune these components and train the retargeter from
scratch with joint trajectory and action supervision:
\begin{equation}
  \begin{aligned}
  \mathcal{L}_{\mathrm{traj}}
    &= \sum_{i,\tau} w_{i,\tau}\,
       \|\vu^{(i)}_{t,\tau}-\vu^{(i)\ast}_{t,\tau}\|_2^2
     + \frac{1}{KT}\sum_{k,\tau}
       \|\vv^{(k)}_{t,\tau}-\vv^{(k)\ast}_{t,\tau}\|_2^2,\\
  \mathcal{L}_{\mathrm{robot}}
    &= \mathcal{L}_{\mathrm{traj}}
     + \lambda\,\big\|\mA_t-\mA_t^{\ast}\big\|_1 .
  \end{aligned}
  \label{eq:loss}
\end{equation}
Here $\ast$ marks ground truth, and $\lambda=1$.
The scene weights $w_{i,\tau}$ sum to one over points and steps and favor motion without excluding static points~\citep{pointworld2026}.

\section{Experiments}
\label{sec:experiments}

In this section, we describe the experimental setup (Section~\ref{sec:setup}) and answer three
questions.
\begingroup
\setlength{\leftmargini}{1.8em}
\begin{enumerate}
  \item[(1)] Does \methodname, pre-trained only on human demonstrations, outperform policies
  pre-trained on robot or video corpora in dexterous and high-precision manipulation?
  (Section~\ref{sec:comparison})
  \item[(2)] Does human-demonstration pre-training transfer to robot control, and does it scale with
  the number of human demonstrations and with model size? (Section~\ref{sec:pretraining})
  \item[(3)] Does 3D world modeling enable more precise dexterous manipulation than 2D world modeling
  or a model without it, and does a 3D encoder help beyond lifted 2D features? (Section~\ref{sec:ablations})
\end{enumerate}
\endgroup

\subsection{Experimental Setup}
\label{sec:setup}

We pre-train \methodname on 1.15M human demonstration episodes from EgoDex~\citep{egodex2026}
and \mbox{VITRA}~\citep{vitra2025}, which provide scene and hand trajectories but no robot actions,
and fine-tune it on the robot demonstrations of each benchmark and of the real robot.

\subsubsection{Benchmarks and Protocols}
\label{sec:benchmarks}
\label{sec:datasets}
\label{sec:eval_protocol}

\textbf{DexJoCo.} DexJoCo~\citep{dexjoco2026} is a simulated benchmark of eleven dexterous tasks,
six single-arm and five bimanual, on Franka Panda arms with 16-DoF Allegro hands. Its tasks require
tool use, such as driving a nail with a hammer, and coordination of two hands. We use its
object-placement setting, where no policy reported with the benchmark exceeds 5.3\% success on
bimanual assembly. We exclude bimanual unlock iPad, since 42 of its 100 demonstrations fail when
replayed in the released environment; 959 of the other 1,000 demonstrations pass the replay check.
We train one policy across the ten tasks (Table~\ref{tab:dexjoco_multitask}) and one per task.
Each task has one fixed instruction, which tells the multi-task policy which task to perform.
Following the official protocol, we run 50 closed-loop episodes per
task on each of three evaluation seeds and report the success rate (SR) averaged over the ten tasks, with its
standard deviation over seeds. Appendix~\ref{app:eval_protocol} details the protocol, and
Appendix~\ref{app:dexjoco} reports the results of the single-task policies.

\textbf{RoboDojo-Precision.} Although \methodname is designed for dexterous hands, its point-trajectory
representation is not tied to one embodiment. We therefore test whether it transfers to parallel-jaw
grippers, using the Precision tasks of RoboDojo~\citep{robodojo2026}, which hinge on precise
placement. Two ARX X5 arms perform these eight tasks, such as inserting a key and fastening screws,
and each task comes with 100 RGBD demonstrations. We train and evaluate
every method on these tasks alone. The tasks are hard; on the benchmark
leaderboard, policies trained on all of its tasks rarely complete them~\citep{robodojo2026}. We therefore also report its
partial-progress score, which separates methods when success is rare.
Appendices~\ref{app:supervision} and~\ref{app:robodojo} detail the setup.

\textbf{Real robot.}
To evaluate \methodname's manipulation performance in the real world, we use an OpenArm humanoid with two 7-DoF arms, two 6-DoF Inspire RH56F1 hands, and a ZED 2i stereo camera, as shown in Fig.~\ref{fig:real_robot_platform}.
We evaluate two tasks: banana pick-and-place (PnP) and square-peg insertion.
Banana PnP evaluates general manipulation ability using a single arm, while square-peg insertion evaluates coordinated dexterous manipulation with both hands.
We collect 54 demonstrations for banana PnP and 64 for square-peg insertion, and evaluate each method over 24 trials per task.
For banana PnP, half of the trials use the banana seen in the demonstrations (in-domain), and the other half use an unseen box whose color resembles the table (out-of-domain). For square-peg insertion, we also report a progress score, the mean number of completed stages per trial out of three, lifting each of the two parts and inserting the peg.

\begin{table}[t]
\centering
\small
\solidrules
\setlength{\tabcolsep}{2pt}
\caption{\textbf{DexJoCo success rate (\%).} One policy across the ten tasks in the object-placement
setting. Subscripts are standard deviations over three seeds, /B marks bimanual tasks, and bold marks
the row best. DP-T and $\pi_{0.5}$ results are from \citet{dexjoco2026}; the others follow the same
protocol.}

\label{tab:dexjoco_multitask}
\begin{tabular}{@{}lrrrrrrO}
\toprule
Task & DP-T & $\pi_{0.5}$ & GR00T N1.6 & Fast-WAM  & PointACT & \methodname \\
\midrule
hammer nail &58.7\sd{5.0} & \textbf{86.7}\sd{3.1} & 82.0\sd{6.0} & 43.3\sd{1.9} & 16.0\sd{8.0} & 85.3\sd{2.3} \\
click mouse &38.7\sd{3.1} & 80.7\sd{3.1} & 67.3\sd{7.6} & 77.3\sd{3.4}  & 37.3\sd{3.1} & \textbf{97.3}\sd{2.3} \\
pick bucket &55.3\sd{7.6} & 83.3\sd{8.1} & 88.7\sd{3.1} & 81.3\sd{6.6}  & 74.7\sd{7.6} & \textbf{98.0}\sd{2.0} \\
pinch tongs &6.0\sd{5.3} & 45.3\sd{6.1} & 28.0\sd{4.0} & 47.3\sd{6.6}  & 26.7\sd{4.2} & \textbf{52.0}\sd{5.3} \\
fold glasses &11.3\sd{5.0} & 42.0\sd{6.0} & 52.0\sd{8.0} & 62.7\sd{3.8} & 20.7\sd{7.6} & \textbf{74.0}\sd{5.3} \\
water plant &60.0\sd{6.9} & 84.0\sd{4.0} & 82.7\sd{3.1} & 42.7\sd{0.9} & 83.3\sd{6.4} & \textbf{86.7}\sd{3.1} \\
hanoi /B &8.0\sd{2.0} & 6.0\sd{0.0} & 34.0\sd{5.3} & 42.0\sd{5.9} & 10.0\sd{2.0} & \textbf{50.0}\sd{9.2} \\
assembly /B &1.3\sd{2.3} & 3.3\sd{2.3} & 6.0\sd{4.0} & 2.7\sd{1.9} & 2.0\sd{0.0} & \textbf{9.3}\sd{4.6} \\
microwave /B &42.7\sd{6.4} & 39.3\sd{13.0} & 78.7\sd{3.1} & \textbf{88.7}\sd{4.1} & 54.0\sd{8.7} & 84.0\sd{4.0} \\
photograph /B &28.0\sd{6.0} & 29.3\sd{1.2} & \textbf{53.3}\sd{8.1} & 40.7\sd{6.2} & 12.0\sd{3.5} & \textbf{53.3}\sd{1.2} \\
\midrule
avg. &31.0\sd{1.7} & 50.0\sd{1.9} & 57.3\sd{1.8} & 52.9\sd{1.5} & 33.7\sd{1.8} & \textbf{69.0}\sd{2.0} \\
\bottomrule
\end{tabular}
\end{table}

\begin{figure}[!t]
\centering
\begin{minipage}[t]{0.24\linewidth}\centering
{\footnotesize hammer nail\strut}\\[1pt]
\includegraphics[width=\linewidth]{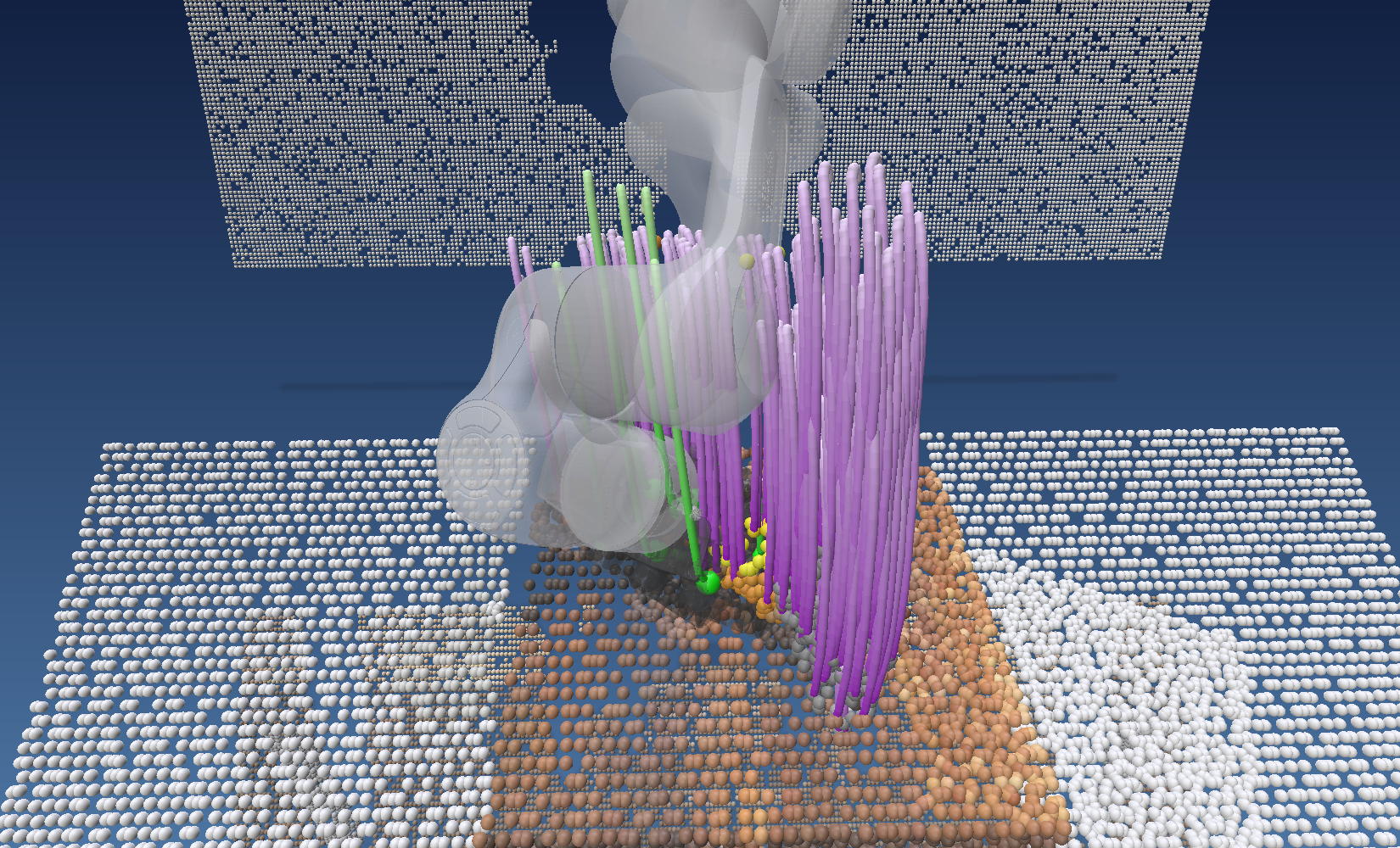}
\end{minipage}\hfill
\begin{minipage}[t]{0.24\linewidth}\centering
{\footnotesize pick bucket\strut}\\[1pt]
\includegraphics[width=\linewidth]{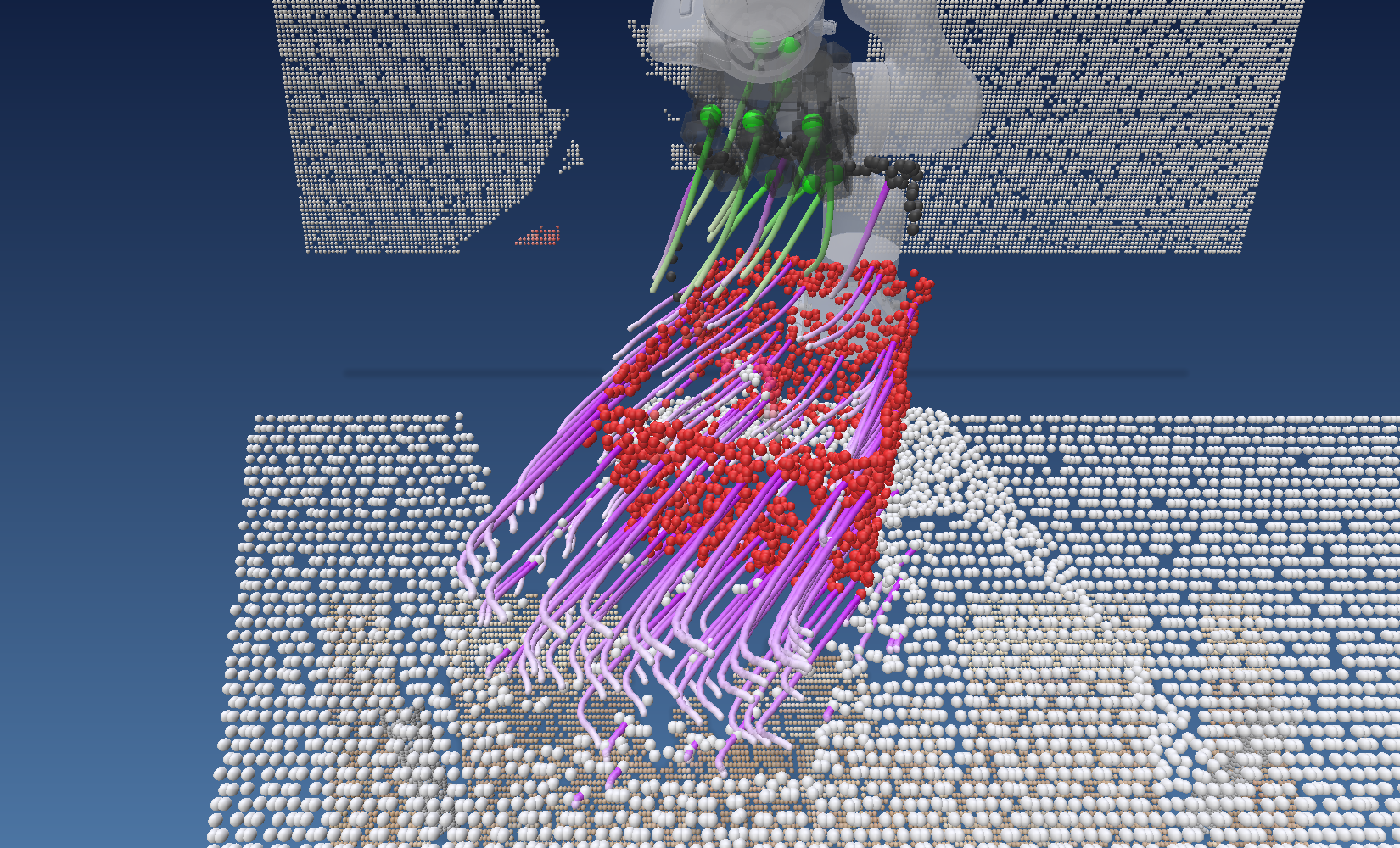}
\end{minipage}\hfill
\begin{minipage}[t]{0.24\linewidth}\centering
{\footnotesize assembly\strut}\\[1pt]
\includegraphics[width=\linewidth]{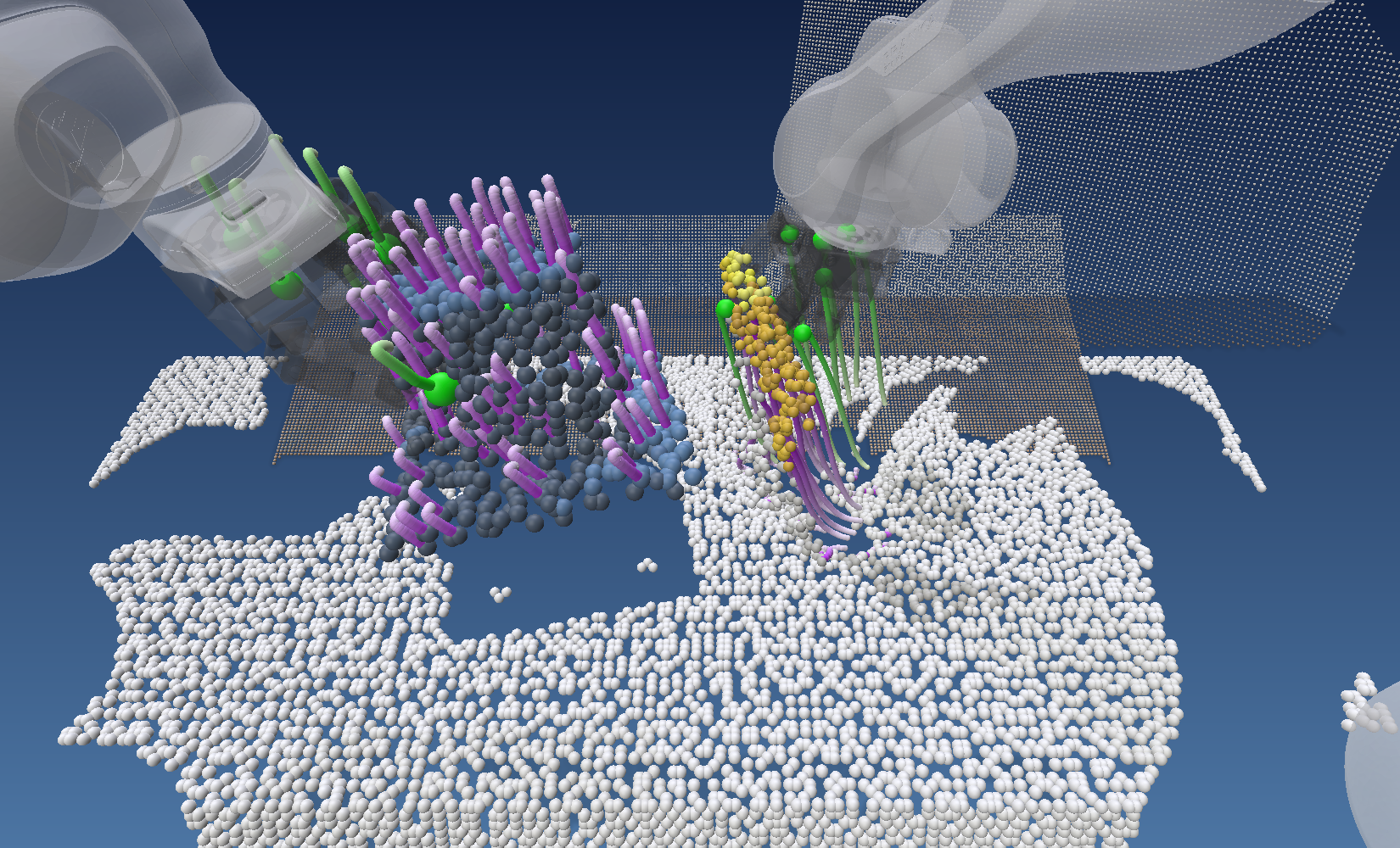}
\end{minipage}\hfill
\begin{minipage}[t]{0.24\linewidth}\centering
{\footnotesize microwave\strut}\\[1pt]
\includegraphics[width=\linewidth]{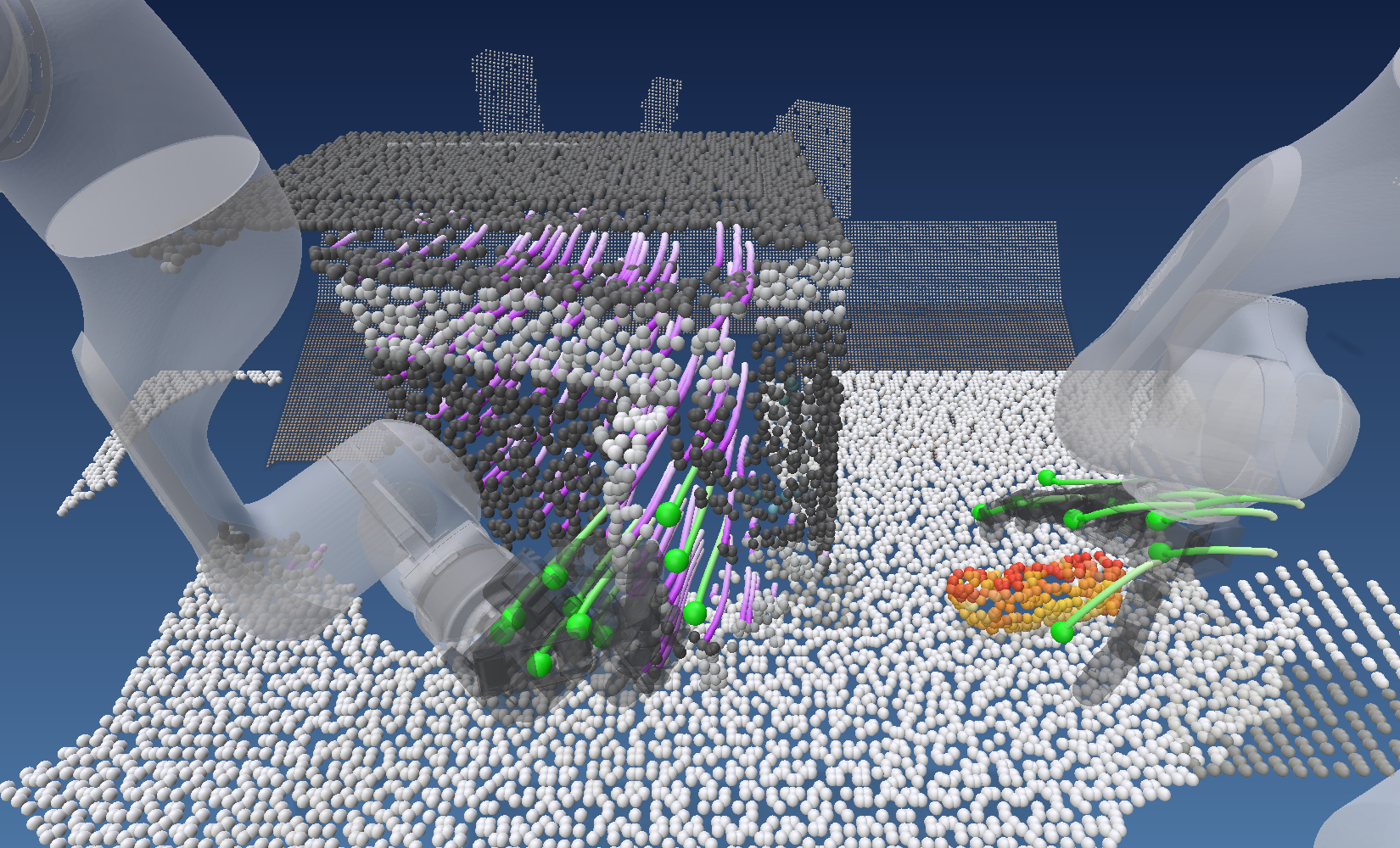}
\end{minipage}
\caption{\textbf{Predicted 3D point trajectories.} Scene ${\color{scenetraj}\vu}$ and hand
${\color{handtraj}\vv}$ trajectories that \methodname predicts on unseen DexJoCo layouts; scene
trajectories are subsampled for visibility.}
\label{fig:qual_traj}
\end{figure}

\subsubsection{Baselines}
\label{sec:baselines}

We take two baselines from DexJoCo~\citep{dexjoco2026} with their published results. DP-T is the
transformer variant of Diffusion Policy~\citep{diffusionpolicy2023}, trained from scratch on images,
and $\pi_{0.5}$~\citep{pi05_2025} is a VLA fine-tuned from a policy pre-trained on robot
demonstrations and web data. We then add three recent families and evaluate them under the same
protocol (Appendices~\ref{app:eval_protocol} and~\ref{app:baselines}). GR00T~N1.6~\citep{groot_n1_2025}
represents VLAs whose pre-training includes large-scale real and simulated robot demonstrations. Fast-WAM~\citep{fastwam2026} represents world action models; it builds on a video generation
backbone and learns robot actions with video-prediction supervision. PointACT~\citep{pointact2026} represents
policies that use depth, adding pre-trained 3D point features to the action expert of a
vision-language policy. Only Fast-WAM forecasts future scenes, and it forecasts
images rather than the 3D motion of the scene and hands.
On the real robot, we compare with GR00T~N1.6 and $\pi_{0.5}$, fine-tuned on the same data.

\subsection{Comparison with Previous Approaches}
\label{sec:comparison}

\begin{table}[!t]
\centering
\footnotesize
\solidrules
\setlength{\tabcolsep}{1.8pt}
\newcommand{\pz}{\phantom{0}}
\caption{\textbf{RoboDojo-Precision results.} Each cell reports the score (left) and success rate in \%
(right); the score measures partial task progress. One policy per method,
one seed, 50 episodes per task.}
\label{tab:robodojo_precision}
\begin{tabular}{lccccccccc}
\toprule
\multirow{2}{*}{Method} & fasten & plug in & insert & pour & play & deposit & insert & build & \multirow{2}{*}{avg.} \\
 & screws & charger & tubes & balls & xylophone & coin & key & tower & \\
\midrule
GR00T N1.6 & 0.4 / 0.0 & 0.0 / 0.0 & 20.4 / \pz8.0 & \pz0.0 / \pz0.0 & 0.0 / 0.0 & \textbf{9.2} / \textbf{4.0} & \textbf{7.8} / 0.0 & \textbf{11.4} / \textbf{6.0} & 6.2 / 2.3 \\
Fast-WAM & 0.0 / 0.0 & 0.0 / 0.0 & \pz0.4 / \pz0.0 & \pz2.0 / \pz2.0 & 0.0 / 0.0 & 0.0 / 0.0 & 0.0 / 0.0 & \pz0.0 / 0.0 & 0.3 / 0.2 \\
PointACT & 0.0 / 0.0 & 0.0 / 0.0 & \pz6.4 / \pz0.0 & \pz0.0 / \pz0.0 & 0.0 / 0.0 & 0.8 / 0.0 & 2.7 / 0.0 & \pz3.2 / 0.0 & 1.6 / 0.0 \\
\rowcolor{defaultrow}
\methodname & \textbf{1.2} / 0.0 & 0.0 / 0.0 & \textbf{32.4} / \textbf{12.0} & \textbf{14.0} / \textbf{14.0} & \textbf{8.0} / \textbf{8.0} & 6.4 / 2.0 & 3.0 / 0.0 & 5.2 / 2.0 & \textbf{8.8} / \textbf{4.8} \\
\bottomrule
\end{tabular}
\end{table}

\begin{figure}[t]
\centering
\definecolor{rolloutsuccess}{HTML}{2CA02C}
\definecolor{rolloutfailure}{HTML}{D62728}
\newlength{\qw}\setlength{\qw}{\dimexpr(\linewidth-35pt)/6\relax}
\newlength{\qh}\setlength{\qh}{0.658\qw}
\newcommand{\qimg}[1]{\includegraphics[width=\qw,height=\qh]{figures/#1}}
\newcommand{\qph}[1]{\includegraphics[width=\qw,height=\qh]{example-image-duck}}
\newcommand{\outcomepanel}[2]{\begin{tikzpicture}[baseline=(img.base)]
  \node[inner sep=0pt] (img) {\qimg{#2}};
  \draw[#1, line width=1.2pt] ([shift={(0.6pt,0.6pt)}]img.south west) rectangle ([shift={(-0.6pt,-0.6pt)}]img.north east);
\end{tikzpicture}}
\newcommand{\outcomeph}[2]{\begin{tikzpicture}[baseline=(img.base)]
  \node[inner sep=0pt] (img) {\qph{#2}};
  \draw[#1, line width=1.2pt] ([shift={(0.6pt,0.6pt)}]img.south west) rectangle ([shift={(-0.6pt,-0.6pt)}]img.north east);
\end{tikzpicture}}
\newcommand{\qrow}[2]{\makebox[11pt][c]{\rotatebox{90}{\makebox[\qh][c]{\scriptsize #1}}}\hspace{2pt}#2}
\newcommand{\qtriple}[3]{#1\hspace{2pt}#2\hspace{2pt}#3}
\newcommand{\qhdr}{\makebox[\qw][c]{\footnotesize $t=0$}\hspace{2pt}%
  \makebox[\qw][c]{\footnotesize $t=t_\mathrm{end}/2$}\hspace{2pt}\makebox[\qw][c]{\footnotesize $t=t_\mathrm{end}$}}
\newcommand{\qgap}{\hspace{8pt}}
\makebox[\linewidth][c]{\makebox[13pt]{}\qhdr\qgap\qhdr}
\\[1pt]
\makebox[\linewidth][c]{\qrow{GR00T~N1.6}{\qtriple{\qimg{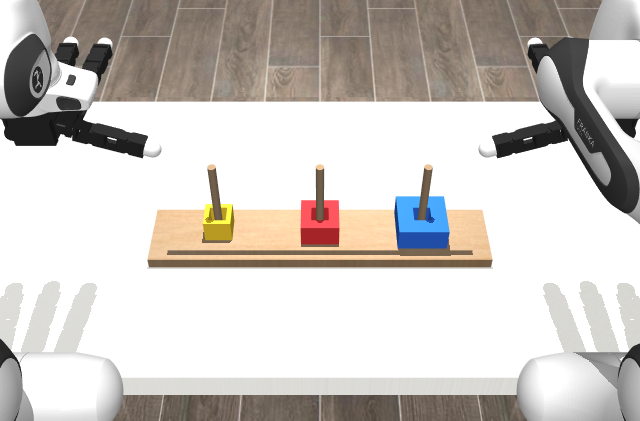}}{\qimg{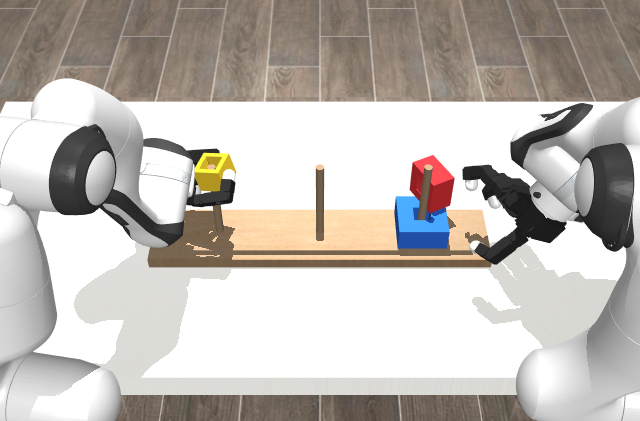}}{\outcomepanel{rolloutfailure}{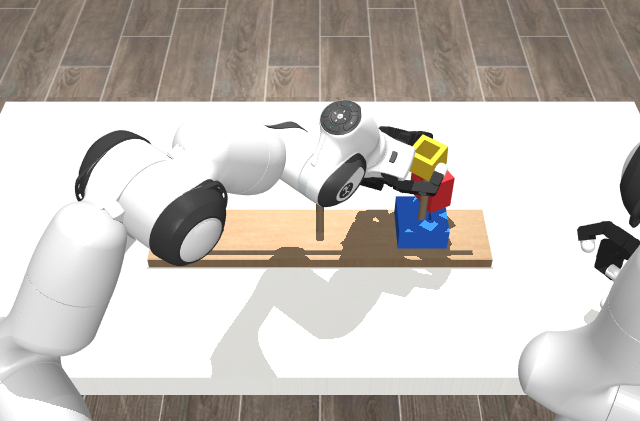}}%
  \qgap\qtriple{\qimg{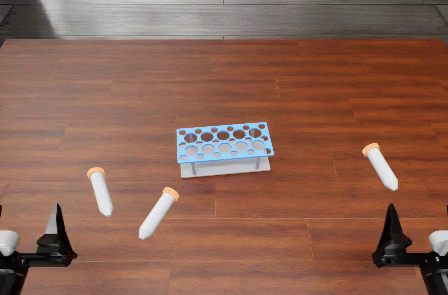}}{\qimg{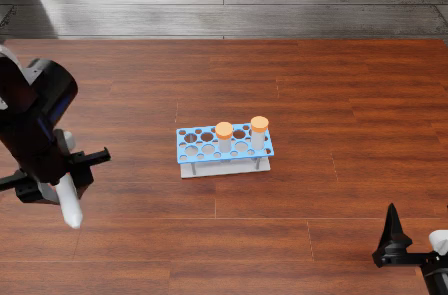}}{\outcomepanel{rolloutfailure}{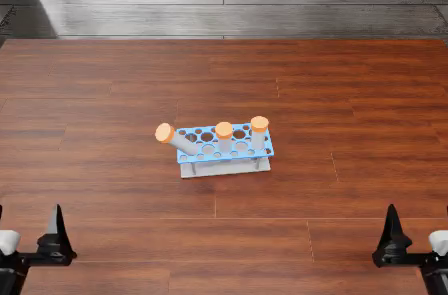}}}}
\\[2pt]
\makebox[\linewidth][c]{\qrow{\methodname}{\qtriple{\qimg{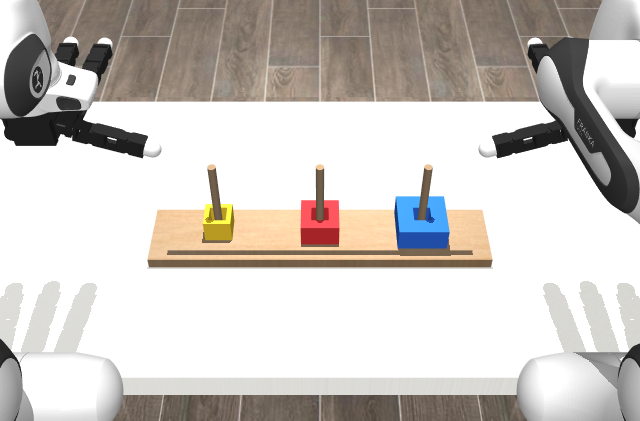}}{\qimg{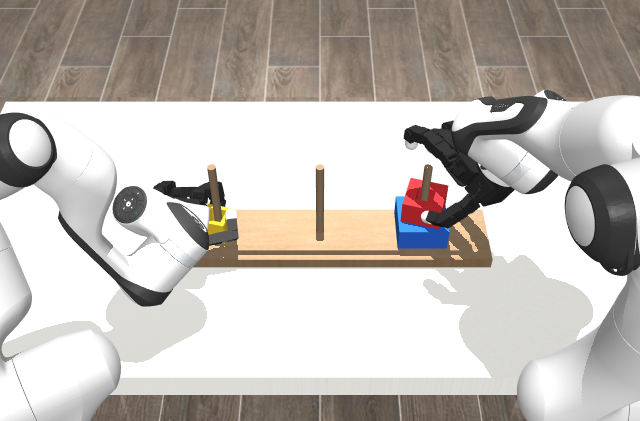}}{\outcomepanel{rolloutsuccess}{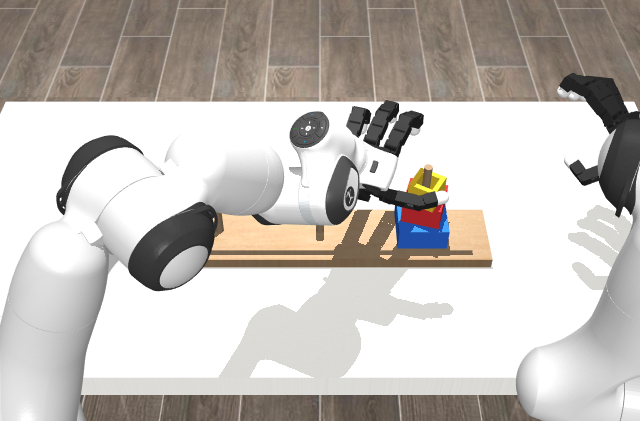}}%
  \qgap\qtriple{\qimg{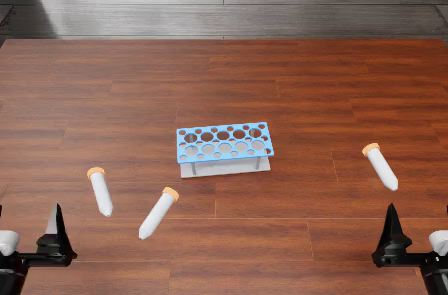}}{\qimg{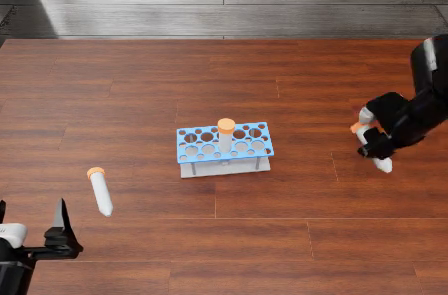}}{\outcomepanel{rolloutsuccess}{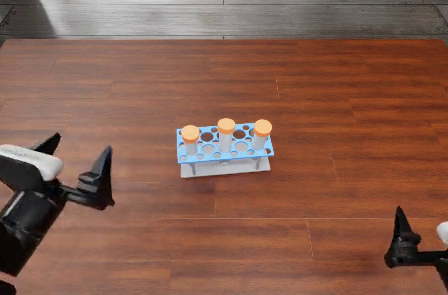}}}}
\\[3pt]
\makebox[\linewidth][c]{\makebox[13pt]{}\makebox[\dimexpr3\qw+4pt][c]{\small (a) DexJoCo (bimanual hanoi)}\qgap%
\makebox[\dimexpr3\qw+4pt][c]{\small (b) RoboDojo-Precision (insert tubes)}}
\caption{\textbf{Qualitative comparison.} Rollouts from one initial state per task, each until
success (\textcolor{rolloutsuccess}{green} border) or failure (\textcolor{rolloutfailure}{red} border)
at its own $t_\mathrm{end}$. \methodname succeeds on both; GR00T~N1.6 times out.}
\label{fig:qual_rollout}
\vspace{-8pt}  %
\end{figure}

\textbf{DexJoCo.} With one policy across the ten tasks, \methodname achieves the highest success rate
on eight tasks and reaches 69.0\% average SR (Table~\ref{tab:dexjoco_multitask}), outperforming the
strongest baseline, GR00T~N1.6, by a large margin of 11.7 percentage points, and $\pi_{0.5}$ by 19.0.
Both pre-train on large-scale robot demonstrations and web data, whereas \methodname uses no
robot demonstrations beyond DexJoCo. \methodname also ranks first with one policy per task,
and its margin over the strongest baseline widens from 5.3 to 22.2
percentage points under full randomization of camera pose, lighting, and texture
(Tables~\ref{tab:dexjoco_main} and~\ref{tab:dexjoco_full} in Appendix~\ref{app:dexjoco}). Figure~\ref{fig:qual_traj} shows the scene and hand trajectories \methodname predicts, and
Fig.~\ref{fig:qual_rollout} shows bimanual hanoi rollouts from one initial state, where \methodname
succeeds and GR00T~N1.6 times out; Appendix~\ref{app:qualitative} shows more rollouts.

\textbf{RoboDojo-Precision.} With no robot data beyond the benchmark's own demonstrations, \methodname reaches an
average score of 8.8 and 4.8\% SR (Table~\ref{tab:robodojo_precision}). Fast-WAM and PointACT also
pre-train on no robot data, only on videos, image-text pairs, or 3D scenes, and stay near zero (0.3, 0.2\% and 1.6, 0.0\%),
even though PointACT observes depth. GR00T~N1.6, whose pre-training includes large-scale robot data
with grippers, reaches 6.2 and 2.3\%, still below \methodname, and leads on three of the eight tasks.
\methodname, pre-trained on human hands alone, thus shows initial transfer to grippers, and depth alone does not
account for its lead over Fast-WAM and PointACT.

\begin{figure}[t]
\centering
{\phantomsubcaption\label{fig:real_robot_platform}\phantomsubcaption\label{fig:real_robot_tasks}}
\newlength{\rpw}\setlength{\rpw}{0.323\linewidth} %
\newlength{\rrb}\setlength{\rrb}{\dimexpr\linewidth-\rpw-6pt\relax}
\definecolor{srgrootgray}{HTML}{B0B0B0}\definecolor{srpigray}{HTML}{6F6F6F}\definecolor{srours}{HTML}{0F4D92}
\newlength{\rtw}\setlength{\rtw}{\dimexpr(\rrb-24pt)/6\relax}
\newlength{\rch}\setlength{\rch}{\dimexpr2\rtw+2pt\relax}
\newlength{\rcw}\setlength{\rcw}{0.315\linewidth} %
\newlength{\rcwb}\setlength{\rcwb}{1.1\rcw}\newlength{\rcwp}\setlength{\rcwp}{0.9\rcw}
\newlength{\rtg}\setlength{\rtg}{12pt} %
\newcommand{\rtask}[1]{\includegraphics[width=\rcwp,height=0.66\rcw]{figures/#1}}
\newcommand{\rhalf}[2]{\begin{tikzpicture}[baseline=(img.south)]
  \node[inner sep=0pt, anchor=south west] (img) {\includegraphics[width=\dimexpr(\rcwb-2pt)/2\relax,height=0.66\rcw]{figures/#1}};
  \node[anchor=north west, inner sep=1.2pt, fill=black, fill opacity=0.55, text opacity=1, text=white,
        font=\fontfamily{phv}\selectfont\tiny, align=left] at ([shift={(1pt,-1pt)}]img.north west) {#2};
\end{tikzpicture}}
\newcommand{\rsr}[2]{\includegraphics[width=#2,height=0.35\rcw]{figures/real_robot_sr_#1}}
\newcommand{\rkey}[1]{\textcolor{#1}{\rule{5pt}{5pt}}}
\newsavebox{\rright}
\sbox{\rright}{\begin{minipage}[b]{\rrb}\centering
\makebox[\linewidth][c]{\makebox[\rcwb][c]{\footnotesize banana pick-and-place}\hspace{\rtg}%
\makebox[\rcwp][c]{\footnotesize square-peg insertion}}\\[1pt]
\makebox[\linewidth][c]{\makebox[\rcwb][c]{\rhalf{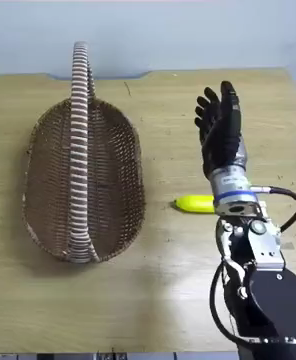}{banana\\(in-domain)}\hspace{2pt}\rhalf{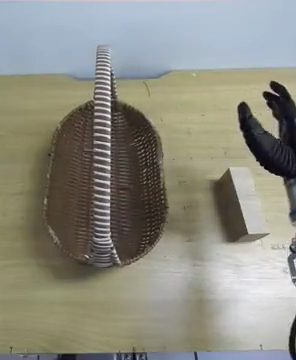}{unseen box\\(out-of-domain)}}\hspace{\rtg}\rtask{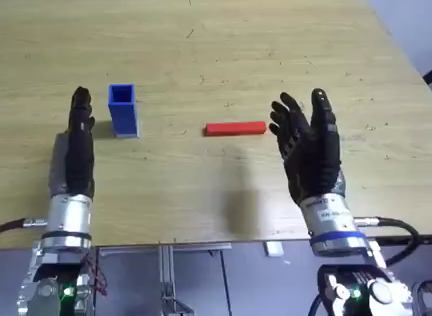}}\\[1pt]
\makebox[\linewidth][c]{\rsr{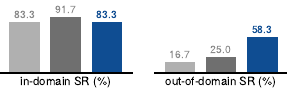}{\rcwb}\hspace{\rtg}\rsr{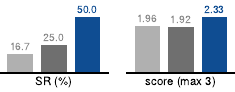}{\rcwp}}\\[0pt]
\makebox[\linewidth][c]{\fontfamily{phv}\selectfont\scriptsize
  \rkey{srgrootgray}~GR00T~N1.6\quad\rkey{srpigray}~$\pi_{0.5}$\quad\rkey{srours}~\methodname}\\[2pt]
{\small (b) Tasks, success rates (SR), and scores over 24 trials}
\end{minipage}}
\begin{minipage}[b]{\rpw}\centering
\begin{tikzpicture}
\node[inner sep=0pt, anchor=south west] (setup) {\includegraphics[height=\dimexpr\ht\rright-12pt\relax,trim=12 0 22 0,clip]{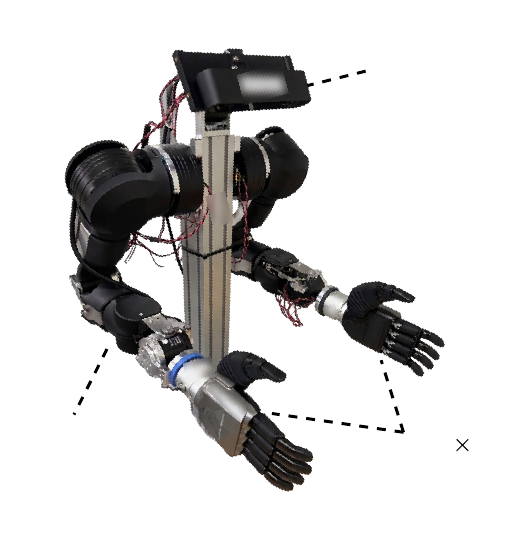}};
\begin{scope}[x={(setup.south east)}, y={(setup.north west)}, font=\fontfamily{phv}\selectfont\scriptsize, align=center, inner sep=0pt]
\fill[white] (0.9475,0.160) rectangle (0.9822,0.188);
\node[anchor=west]  at (0.7683,0.867) {ZED 2i\\(Stereo)};
\node[anchor=north] at (0.1078,0.217) {OpenArm\\(7 DoF)};
\node[anchor=north] at (0.8342,0.182) {RH56F1\\($\times$2)};
\end{scope}
\end{tikzpicture}\\[1pt]
{\small (a) Platform}
\end{minipage}\hspace{6pt}%
\usebox{\rright}
\caption{\textbf{Real-robot evaluation.} (a) The OpenArm platform. (b) The two tasks, one single-arm
and one bimanual, with the success rate and score of each method.}
\label{fig:real_robot}
\vspace{-8pt}  %
\end{figure}

\textbf{Real robot.}\label{sec:realrobot}
Figure~\ref{fig:real_robot_tasks} summarizes the real-robot results. \methodname outperforms both
GR00T~N1.6 and $\pi_{0.5}$ over all trials of each task. On banana PnP, it performs comparably with
the in-domain banana but reaches 58.3\% SR with the out-of-domain unseen box, against 16.7\% and
25.0\%. Since the box resembles the table in color, the RGB-based baselines likely struggle to find
it, while \methodname can rely on its 3D shape. On square-peg insertion, \methodname reaches 50.0\% SR
against 16.7\% and 25.0\%. All three methods usually lift both parts, but the baselines often fail
at the insertion, which needs precise bimanual placement; \methodname inserts in 12 of 21 such
trials, against 4 of 19 and 6 of 18. Figure~\ref{fig:app_qual_real} in Appendix~\ref{app:qualitative}
shows example rollouts.

\subsection{Human-Demonstration Pre-Training and Scaling}
\label{sec:analysis}
\label{sec:pretraining}

\begin{figure}[t]
\captionsetup{justification=raggedright,singlelinecheck=false}
\begin{minipage}[t]{0.315\linewidth}
\centering
\includegraphics[width=\linewidth]{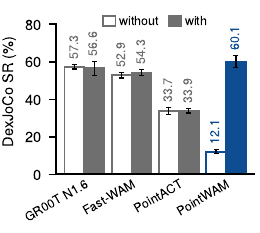}
\caption{\textbf{Pre-training gain.} Average SR~(\%) with and without
EgoDex pre-training.}
\label{fig:pretraining_gain}
\end{minipage}\hfill
\begin{minipage}[t]{0.315\linewidth}
\centering
\includegraphics[width=\linewidth]{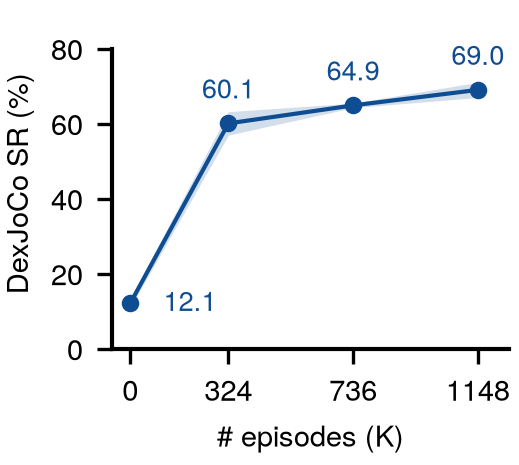}
\caption{\textbf{Data scaling.} Each point plots average SR~(\%) versus
pre-training episodes.}
\label{fig:scaling_data}
\end{minipage}\hfill
\begin{minipage}[t]{0.315\linewidth}
\centering
\includegraphics[width=\linewidth]{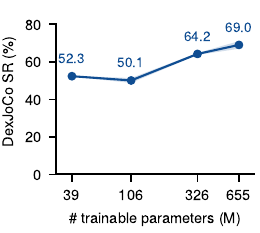}
\caption{\textbf{Model scaling.} Each point plots average SR~(\%) versus
trainable parameters.}
\label{fig:scaling_model}
\end{minipage}
\vspace{-10pt}  %
\end{figure}

\textbf{Transfer from human demonstration videos.} We first measure how much EgoDex~\citep{egodex2026}
pre-training improves each method (Fig.~\ref{fig:pretraining_gain}; whiskers show $\pm1$ standard
deviation over three seeds), fixing Volt-H and 60k fine-tuning iterations for \methodname. The
baselines already build on large \emph{pre-trained} models; GR00T~N1.6 inherits vision-language and
robot-policy pre-training~\citep{groot_n1_2025}, and Fast-WAM and PointACT build on pre-trained
Wan2.2~\citep{wan2025} video and Qwen2.5-VL~\citep{qwen25vl2025} backbones~\citep{fastwam2026,pointact2026}.
\methodname instead trains its backbone \emph{from scratch}, keeping only its frozen encoders
pre-trained, and hence starts from a lower SR than the baselines.

\methodname nevertheless benefits far more from EgoDex pre-training. Its SR rises nearly fivefold,
from 12.1\% to 60.1\%, whereas the baselines change by at most 1.4 percentage points (GR00T~N1.6
from 57.3\% to 56.6\%, Fast-WAM from 52.9\% to 54.3\%, and PointACT from 33.7\% to 33.9\%). Even
Fast-WAM, which also forecasts the future but as video frames, gains little. These results suggest
that the gain comes from the shared 3D representation, in which human and robot demonstrations take
the same scene and hand point trajectories, rather than from world modeling alone.

\textbf{Data scaling.}\label{sec:scaling} We next vary the amount of human demonstrations with the
Volt-H backbone and 60k fine-tuning iterations fixed, adding half and all of VITRA~\citep{vitra2025}
to the 324K EgoDex episodes (Fig.~\ref{fig:scaling_data}; shading shows $\pm1$ standard deviation
over three seeds). SR rises from 60.1\% with EgoDex alone to 64.9\% with 736K episodes and 69.0\% with the full
1.15M episodes. EgoDex alone already surpasses GR00T~N1.6 (57.3\%), and the full corpus adds
8.9 percentage points.

\textbf{Model scaling.} We then vary the trainable capacity of the backbone with the full EgoDex and VITRA corpus and 60k fine-tuning iterations fixed
(Fig.~\ref{fig:scaling_model}; shading shows $\pm1$ standard deviation over three seeds).
SR stays near 50\% for Volt-S (39M, 52.3\%) and Volt-B (106M, 50.1\%) and rises to 64.2\% for
Volt-L (326M) and 69.0\% for Volt-H (655M). Appendix~\ref{app:training} lists the configurations.

\subsection{Component Analysis}
\label{sec:ablations}

\begin{table}[t]
\centering
\caption{\textbf{Component ablations.} Average SR (\%) of a Volt-L12 model pre-trained on EgoDex,
varying one setting per panel (Section~\ref{sec:ablations}). Shading marks the reference, and
subscripts are standard deviations over three seeds of 50 episodes per task.}
\label{tab:design_ablation}

\abcell{a}{vision encoder}{case & SR (\%)}{%
DINOv3 & 59.0\sd{1.3} \\
\dfltrow{Mosaic3D}{64.4\sd{1.7}}}\hspace{0.3in}%
\abcell{b}{world modeling}{case & SR (\%)}{%
without & 53.5\sd{2.7} \\
2D frame & 60.1\sd{1.5} \\
\dfltrow{3D points}{64.4\sd{1.7}}}\hspace{0.3in}%
\abcell[lrc][3]{c}{hand representation}{case & points & SR (\%)}{%
\dfltrowiii{keypoints}{10}{64.4\sd{1.7}}
surface points & 10 & 43.6\sd{0.8} \\
surface points & 512 & 44.6\sd{0.9} \\}\hspace{0.3in}%
\abcell{d}{voxel size}{size & SR (\%)}{%
0.5\,cm & 59.9\sd{1.5} \\
\dfltrow{1\,cm}{64.4\sd{1.7}}
2\,cm & 52.2\sd{1.2} \\}
\vspace{-10pt}  %
\end{table}

We change one setting at a time from a common reference, a Volt-L12 backbone, named for halving the
depth of Volt-L to 12 layers, pre-trained on EgoDex and fine-tuned for 60k iterations
(Table~\ref{tab:design_ablation}).
Since the ablations require many training runs, the reference is smaller and uses fewer human
demonstration videos than the benchmark model, so its SR is lower than in
Table~\ref{tab:dexjoco_multitask}. Appendix~\ref{app:training} gives the training settings, and
Appendix~\ref{app:additional_ablations} further varies the patch size and encoder fine-tuning.

\textbf{Vision encoder (a).} The 3D encoder Mosaic3D, which computes features on the point cloud,
reaches 64.4\%. Lifting \mbox{DINOv3}~\citep{dinov32025} image features onto the points, as in
PointWorld~\citep{pointworld2026}, lowers SR to 59.0\%. Lifted features carry 3D positions but are
computed from images. The gap suggests that encoding the 3D structure of the scene in the features
themselves, not only in their positions, matters for dexterous manipulation.

\textbf{World modeling (b).} Removing scene-trajectory supervision from both pre-training and
fine-tuning, with the depth input, 3D backbone, and hand head unchanged, lowers SR from 64.4\% to
53.5\%. Forecasting the scene thus improves success beyond observing it in 3D, and
Fig.~\ref{fig:qual_traj} shows the predicted scene moving with the hands. The 2D frame row
uses the DINOv3 encoder of (a) and forecasts the DINOv3 latents of the frame at $t+T$ through
learnable queries that read the backbone's output tokens, as in
FLARE~\citep{flare2025} but with DINOv3 rather than SigLIP-2 latents, so that its input and its
target share one feature space. It reaches 60.1\%, 6.6 percentage points above the model without world modeling but 4.3 below the
reference. Forecasting the 3D trajectories of scene points from 3D features thus outperforms
forecasting future image latents from image features.

\textbf{Hand representation and voxel size (c--d).} In (c), anatomical keypoints reach 64.4\%,
whereas depth-sampled hand surface points reach 43.6\% with 10 points per hand and 44.6\% with 512.
Keypoints likely help because they correspond between human and robot hands
(Section~\ref{sec:formulation}). In (d), 1\,cm voxels reach 64.4\%, above both finer 0.5\,cm voxels
(59.9\%) and coarser 2\,cm voxels (52.2\%). Coarser voxels may blur the contact geometry, while
finer ones add little detail beyond the $224\times224$ depth input. We therefore use 1\,cm voxels in
all experiments, including Table~\ref{tab:dexjoco_multitask}.

\WFclear

\Needspace{10\baselineskip}
\section{Conclusion}
\label{sec:conclusion}

We have introduced \methodname, a 3D world action model that decomposes the world into a scene and hands.
It jointly forecasts both as 3D point trajectories in a shared space-time coordinate frame and
retargets the forecast hand motion to robot actions. Since human videos provide the same
trajectories, \methodname pre-trains on them without selecting task-specific points, raising average
DexJoCo success from 12.1\% to 69.0\% and surpassing the strongest baseline by 11.7 percentage points.
Observing the scene in 3D per se is not enough; forecasting the scene adds 10.9 percentage points
over forecasting the hands alone. These results suggest that a shared 3D point-trajectory
representation lets robots learn dexterous manipulation largely from human videos. Since
\methodname sees the world only through points, it can miss details finer than the point spacing,
such as printed text, and surfaces that depth sensing cannot capture, such as transparent objects.
Future work could extend what each point carries, from finer appearance to physical properties such
as mass and friction, which govern how objects move under contact.

\clearpage
%

\bibliography{main}
\bibliographystyle{iclr2027_conference}

\clearpage
\appendix

\section{Extended Related Work}
\label{app:related}

In this section, we extend Section~\ref{sec:related} to further world action models, 3D
representations for robot policies, and other ways of learning from human demonstration videos.

\paragraph{Future targets of world action models.}
Beyond the dense 3D scene motion forecast by EgoWAM and PointWorld, world action models differ in
what future they forecast and what they condition it on. Several forecast geometric futures, such as
object poses in DyWA~\citep{dywa2025} and depth in GAM~\citep{gam2026}. 3D-VLA~\citep{threedvla2024}
forecasts goal RGBD images or point clouds, and WSA$_1$~\citep{wsa2026} forecasts latent 3D scene
states. Others keep an image future. Fast-WAM~\citep{fastwam2026} and Cosmos~3~\citep{cosmos3} build
on video generation backbones, 4D-WAM~\citep{fourdwam2026} aligns its features with 3D trajectory
fields during training, and FLARE~\citep{flare2025} aligns future tokens of its policy with image
features of a future frame. X-WAM~\citep{xwam2026} forecasts multi-view RGB-D video
alongside actions. In each of these, the forecast future takes a different form from the
motion that produces the action, whereas \methodname expresses the scene and the hands with the same
3D point trajectories and retargets only the hands to actions. Dexterous World Model~\citep{dwm2026}
conditions its interaction video on given hand trajectories, much as PointWorld conditions on
future robot trajectories, so it also requires a supplied hand trajectory rather than forecasting one itself for robot control.

\paragraph{3D representations for robot policies.}
\methodname observes the scene as a point cloud, which gives a policy direct access to 3D geometry.
DexPoint~\citep{dexpoint2022} and DexArt~\citep{dexart2023} combine observed geometry with
kinematics-derived point clouds for dexterous reinforcement learning, and DP3~\citep{dp32024}
learns imitation policies from encoded point clouds. Even the image-based
SpatialVLA~\citep{spatialvla2025} augments its features with 3D positions from estimated depth.
In all of these, geometry serves only as input, and the control objective alone trains the
backbone. CordViP~\citep{cordvip2025} adds contact and hand--arm correspondence objectives.
PointACT~\citep{pointact2026} keeps a 2D vision--language backbone and admits pre-trained 3D point
features only in its action expert, through multi-scale point--action attention, so geometry
conditions the action but never the backbone's representation. \methodname also builds on
pre-trained 3D features but supervises the backbone with future hand and scene trajectories, which
human demonstration videos supply without recorded robot actions.

\paragraph{Other ways of learning from human demonstration videos.}
Beyond the point-based policies in Section~\ref{sec:related}, methods differ in what they transfer
from the video to the robot. Retargeting the human hand
pose~\citep{dexmv2022} transfers actions directly but sets no scene-forecasting objective, and neither
do the hand and object poses of FlowHOI~\citep{flowhoi2026}. DexWM~\citep{dexwm2025} learns a latent world model from
egocentric human videos with hand keypoints as actions, so it forecasts the effect of given hand
motion rather than the motion itself, and HarmoHOI~\citep{harmohoi2026} co-generates hand--object
interaction videos and 3D point tracks for synthesis rather than robot control. Track-based methods~\citep{atm2024,track2act2024}
forecast the future image positions of arbitrary points, but the tracks and the features behind
them stay in image space. General Flow~\citep{generalflow2024} forecasts 3D flow only at query
points sampled on the manipulated object, MolmoMotion~\citep{molmomotion2026} forecasts 3D
trajectories of query points on an object of interest, and $\mu_0$~\citep{mu02026} forecasts 3D
traces of interaction keypoints selected with a vision-language model. In all three, something outside the model
decides which scene points it supervises. \methodname instead encodes
the full point cloud and supervises the trajectory of every observed point and of the hand, learned
from human demonstration videos without robot actions.

\section{Implementation Details}
\label{app:implementation}

Appendix~\ref{app:human_preprocessing} describes how we convert human demonstrations into
point trajectories, and Appendix~\ref{app:supervision} does the same for DexJoCo and ends with the
changes for RoboDojo-Precision. The remaining subsections give the model and training settings.

\subsection{Human-Demonstration Preprocessing}
\label{app:human_preprocessing}

Human videos record neither robot states nor simulator states, so we estimate hand and scene
trajectories from the videos themselves. EgoDex~\citep{egodex2026} annotates an $SE(3)$ transform
for each of 25 joints per hand, and VITRA~\citep{vitra2025} reconstructs 21 joints for each visible
hand. We keep the joint positions and map them to the hand keypoints described below. When a hand
is missing, we mark its keypoints invalid instead of imputing them.

\textbf{Hand keypoints.} We turn each human hand into the same ten keypoints as the robot hand
(Appendix~\ref{app:supervision}). Human demonstrations are first rotated, with their scenes, into a
z-up frame like the robot-base frame of robot demonstrations. At each time $t$, we build a palm frame $(\vo_t,\mB_t)$ on
the human hand, with its origin at the middle knuckle and its axes along the wrist-to-knuckle
direction, along the knuckle row, and their cross product. Indexing the ten keypoints of one hand by
$k$, the six palm keypoints overlay the robot palm on the human palm, placing the Allegro palm
points at their fixed offsets $\vd^{(k)}$ from the robot's palm in this frame. The four
fingertip keypoints are the tracked tips $\vf^{(k)}_t$ of the thumb, index, middle, and ring fingers:
\begin{equation}
  \vq^{(k)}_t = \vo_t + \mB_t\,\vd^{(k)} \quad (k=1,\ldots,6),
  \qquad
  \vq^{(k)}_t = \vf^{(k)}_t \quad (k=7,\ldots,10).
  \label{eq:handmap}
\end{equation}
The little finger is omitted because the Allegro hand has no counterpart to it.

\textbf{Clips.} We split each episode into 1\,s clips of 11 frames at 10\,Hz that start every
0.5\,s, and discard clips whose summed wrist displacement is below 2\,cm. We linearly resample each
clip to 31 frames, so that its future steps match the horizon $T=30$ of the DexJoCo demonstrations.
EgoDex contributes 5.2M clips from 324K episodes, and VITRA contributes 824{,}157 shorter episodes.

\textbf{Depth.} VGGT-$\Omega$~\citep{wang2026vggtomega} predicts per-frame depth only up to scale.
We recover the metric scale from the hands, which ARKit annotates in EgoDex and VITRA reconstructs.
For each chunk of frames, we project the hand joints into the images, discard depth ratios
$z_{\text{GT}}/z_{\text{pred}}$ more than a factor of two from their median, and take the median of
the remaining ratios as the scale.

\textbf{Scene tracks.} We resize frames to $640\times360$, place queries on every second pixel of
each clip's first frame, and track them with CoTracker3~\citep{karaev2025cotracker3} in offline
mode. Backward tracking is unnecessary, since every query starts at the first frame. Each track
reads the scaled depth at its pixels and is lifted to the world frame with the camera intrinsics
and poses. We keep points that are visible with valid depth in the first frame, deduplicate them on
a 1.5\,cm grid, and subsample at most 12{,}000. Since queries restart in every clip, point
correspondence holds only within a clip.

\subsection{Robot-Demonstration Preprocessing}
\label{app:supervision}

\textbf{Point cloud and scene trajectories.} In DexJoCo, robot fine-tuning and evaluation use $224\times224$
depth images, which we unproject with the camera calibration, reduce to one point per 1\,cm voxel,
and crop to a box around the robot base, 1\,m to each side and from 0.2\,m below to 1\,m above it.
The box adapts the bounds filter of PointWorld~\citep{pointworld2026}, a cube about the scene
center, to exclude the floor and walls. Since the DexJoCo dataset records object poses only at
reset, we replay each demonstration to recover them over time. For each successful replay, simulator
segmentation removes robot points from the initial point cloud and assigns every remaining point,
background included, to an object or rigid part, whose rigid motion defines the point's trajectory.

\textbf{Robot hand keypoints.} Each robot hand has ten keypoints, six on the palm and four on the
fingertips. The fingertip keypoints lie on the contact pads, since points on the distal joint axes
would miss the motion of four finger joints during flexion. Separate kinematic chains compute the
right and left hands, which record their joints in different orders, and each hand's measured pose
places its keypoints in the scene frame without mirroring one hand onto the other. Bimanual inputs
list the right hand before the left and give each hand its own ID.

\textbf{RoboDojo-Precision.} Its two ARX X5 arms carry parallel-jaw grippers, so we keep the
pre-trained interface and change only what the embodiment requires. Following Point Policy~\citep{pointpolicy2025}, which
places robot points by a rigid transform about the gripper pose, each arm again has ten keypoints,
the six palm offsets placed on the gripper frame from forward kinematics and the two fingertips,
with the last point repeated to fill the ten slots. The point cloud fuses the depth of
the head and two wrist cameras and drops points within 5\,cm of the arm and finger segments. The
action replaces the sixteen finger joints with one gripper command, so $d_a=7$ per arm, and the
30-step chunk spans 1.2\,s at the benchmark's 25\,Hz. Since the released demonstrations carry no
object poses or segmentation, scene trajectories cannot be reconstructed, and fine-tuning supervises
only the hand keypoints and actions while the scene head stays frozen.

\subsection{3D Trajectory Forecaster}

\textbf{Tokenization.} We build on the Volt~\citep{yilmaz2026volt} implementation in
WarpConvNet~\citep{warpconvnet2025}. Where Volt projects the raw voxel features of each patch, this
implementation first computes a feature $\vg^{(c)}\in\sR^{d_{\mathrm{stem}}}$ for every occupied
voxel $c$ with a sparse-convolutional stem. The stem applies a $3^3$ sparse convolution to
$d_{\mathrm{stem}}=64$ channels, LayerNorm~\citep{ba2016layernorm}, GELU~\citep{hendrycks2016gelu},
and two residual blocks. The token of patch $j$ concatenates the stem features of its $K_v^3$ voxels
in a fixed spatial order, with zeros for empty voxels, and projects them to width $d$:
\begin{equation}
  \vz^{v}_j = \mW_{v}
  [\vg^{(c_1)};\ldots;\vg^{(c_{K_v^3})}] + \vb_{v}
  \in\sR^d.
  \label{eq:token}
\end{equation}
Each patch spans $K_v=5$ voxels, or 5\,cm, per side, and the fixed order preserves where each voxel
sits inside the patch, which pooling would discard.

\textbf{Backbone.} The backbone is Volt-H, with 32 layers, 20 heads, and width $d=1280$. The $L$
instruction features from the frozen Recap-CLIP text encoder~\citep{li2024recaption} are projected
to width $d$ and precede the $M$ vision tokens in Eq.~\eqref{eq:joint_attention}. Axial 3D rotary
position embeddings~\citep{su2024roformer,yilmaz2026volt} encode the patch coordinates of vision
tokens. Instruction tokens receive no rotation, since the text encoder already encodes word order.
Attention is bidirectional and stays within one observation, and only the output vision tokens are
decoded.

\textbf{Trajectory decoder.} Before the Detokenizer of Eq.~\eqref{eq:trajectory_decoder} expands a
token, the token passes through LayerNorm, GELU, a linear layer, LayerNorm, and GELU. The saved
mapping $\mathcal{I}_t$ then selects each point's feature from the $K_v^3$ expanded voxel features.
The stem features $\mS_t\in\sR^{(N+K)\times d_{\mathrm{stem}}}$ give each scene point or hand
keypoint the feature $\vg^{(c)}$ of its voxel. The fusion MLP consists of a linear layer, LayerNorm,
GELU, and a zero-initialized linear layer, so training starts from the detokenized features alone.
A final LayerNorm and GELU applied to the residual sum give the per-point features $\mF_t$
that the scene and hand heads of Eq.~\eqref{eq:trajectory_heads} read to forecast trajectories.

\subsection{Action Retargeter}

The retargeter of Eq.~\eqref{eq:action_retargeter} stacks two decoder blocks with four attention
heads. $\operatorname{MLP}_{\mathrm{disp}}$ and $\operatorname{MLP}_{\mathrm{state}}$ each have two
linear layers with a SiLU between them, and $\operatorname{MLP}_{\mathrm{disp}}$ reads the $K$ rows
of $\tV_{t,\tau}$ as one vector in a fixed order, which preserves hand and keypoint identities.
Each block applies LayerNorm before its self-attention, cross-attention, and feed-forward layers.
Self-attention among the action queries is unmasked, so the decoder predicts the whole action chunk
in parallel rather than one step at a time. With $\mH_t\in\sR^{T\times d_r}$ the output queries of
the last block, the final linear layer follows a LayerNorm:
\begin{equation}
  \mA_t = \operatorname{LayerNorm}(\mH_t)\,\mW_{a} + \vb_{a}.
  \label{eq:action_retargeter_full}
\end{equation}
The retargeter learns the map from robot demonstrations instead of solving inverse kinematics, so
it enforces no explicit joint limits or collision constraints. Its targets are recorded actions,
which already respect joint limits and avoid collisions since a robot executed them.

\subsection{Normalization and Loss Weighting}
\label{app:normalization}

\textbf{Trajectory normalization.} The heads of Eq.~\eqref{eq:trajectory_heads} forecast
standardized displacements. We standardize scene and hand displacements separately, with
coordinate-wise means $\vmu^{\mathrm{scene}}_\tau,\vmu^{\mathrm{hand}}_\tau\in\sR^3$ and standard
deviations $\vsigma^{\mathrm{scene}}_\tau,\vsigma^{\mathrm{hand}}_\tau\in\sR^3$ for each horizon
step, and fine-tuning keeps the statistics of pre-training. With $\bar{\vu}^{(i)}_{t,\tau}$ and
$\bar{\vv}^{(k)}_{t,\tau}$ the entries of $\tU_t$ and $\tV_t$, we recover the displacements
of Eq.~\eqref{eq:motion_target} as
\begin{equation}
  \vu^{(i)}_{t,\tau} = \vmu^{\mathrm{scene}}_\tau + \vsigma^{\mathrm{scene}}_\tau\odot\bar{\vu}^{(i)}_{t,\tau},
  \qquad
  \vv^{(k)}_{t,\tau} = \vmu^{\mathrm{hand}}_\tau + \vsigma^{\mathrm{hand}}_\tau\odot\bar{\vv}^{(k)}_{t,\tau},
  \label{eq:model}
\end{equation}
where $\odot$ denotes coordinate-wise multiplication. Adding them to $\vp^{(i)}_t$ and $\vq^{(k)}_t$
gives the forecast positions. Before standardization, we clip every coordinate of the scene
displacements to the range $[-0.5,0.5]$\,m, and the losses of Eq.~\eqref{eq:loss} compare
standardized displacements.

\textbf{Action normalization.} Following the Franka--Allegro interface of
DexJoCo~\citep{dexjoco2026}, each arm has three end-effector position coordinates, three rotation
coordinates, and sixteen finger-joint targets, so $d_a=22$ per arm. We express each target relative
to the robot state $\vs_t$, as differences for positions and finger joints and as a rotation vector
for orientation, then standardize each coordinate per horizon step with statistics of the robot
demonstrations. The target $\mA_t^{\ast}$ of Eq.~\eqref{eq:loss} is a recorded chunk after this map,
which we invert at inference. Single-arm and bimanual embodiments have
separate retargeters and action statistics but share the trajectory forecaster, whose per-point
outputs do not depend on the layout of the action space.

\textbf{Loss weighting.} In Eq.~\eqref{eq:loss}, a hand keypoint is valid when its hand is present
and fills a real slot of the bimanual layout rather than padding, and the hand loss averages over valid keypoint--step pairs.
Following PointWorld~\citep{pointworld2026}, the scene weights exclude points without a valid target
and softly emphasize motion between adjacent frames. Their unnormalized values are
$\operatorname{sigmoid}(5(\delta_{i,\tau}/0.005-1))$, with
$\delta_{i,\tau}=\|\vp^{(i)}_{t+\tau}-\vp^{(i)}_{t+\tau-1}\|_2$ in meters, so a static point keeps a
small positive weight; we normalize them over valid point--step pairs in the batch. Each squared
error is divided by 3 and the action error by $Td_a$, so every term is a per-coordinate mean. Hand
and action errors average over valid batch elements, and all loss coefficients equal one.

\section{Experimental Details}
\label{app:training}

\subsection{Training Settings}

\begin{table}[t]
\centering
\small
\setlength{\tabcolsep}{5pt}
\caption{\textbf{Training settings for human-demonstration pre-training and DexJoCo fine-tuning.}}
\label{tab:training_details}
\begin{tabular}{@{}lll@{}}
\toprule
 & Pre-training & Fine-tuning \\
\midrule
Data & EgoDex + VITRA (1.15M episodes) & DexJoCo (959 demos) \\
Iterations & 60k & 60k \\
Global batch size & 192 & 192 \\
GPUs & 4 NVIDIA B200 & 4 NVIDIA B200 \\
Optimizer & AdamW~\citep{loshchilov2019adamw} & AdamW \\
Learning rate & $10^{-4}$ & $10^{-4}$ \\
Learning-rate schedule & constant & constant \\
Weight decay & 0.01 & 0.01 \\
Warm-up & none & none \\
Gradient-norm clipping & 5.0 & 5.0 \\
Stochastic depth~\citep{huang2016stochasticdepth} & 0 to 0.3 over depth & 0 to 0.3 over depth \\
Precision & bfloat16 mixed & bfloat16 mixed \\
Window length & 31 frames (1\,s of video) & 31 frames (0.6\,s simulated) \\
Voxel size & 1\,cm & 1\,cm \\
Patch size $K_v$ & 5 & 5 \\
Maximum scene points & 12,000 & unrestricted \\
Hand keypoints & 20 & 10 or 20 \\
Trainable / frozen parameters & 651M / 592M & 655M / 592M \\
Frozen encoders & Mosaic3D, Recap-CLIP text & Mosaic3D, Recap-CLIP text \\
Retargeter & not used & trained \\
Initialization & random & pre-training checkpoint (60k) \\
Training time & 34\,h & 19\,h \\
\bottomrule
\end{tabular}

\end{table}

\textbf{Main model.} Table~\ref{tab:training_details} lists the settings of the Volt-H multi-task
model in Table~\ref{tab:dexjoco_multitask}. We pre-train it for 60k iterations on EgoDex and VITRA
(1,148,324 episodes in total) and fine-tune it for 60k iterations on DexJoCo, with the 3D and text
encoders frozen in both stages.

\textbf{Component ablations.} Ablations use Volt-L12, a backbone with width 1024, 12 layers, and 16
heads, which trains 175M parameters. We pre-train it on EgoDex for 40k iterations and fine-tune it
for 60k, starting from the pre-training snapshot with the lowest held-out hand-trajectory error
(35k for the reference). Every variant, surface points included, changes both pre-training and fine-tuning.

\textbf{Model scaling.} The model-scaling study varies backbone width and depth while keeping the
other components fixed. The Volt-S, Volt-B, Volt-L, and Volt-H configurations in
Fig.~\ref{fig:scaling_model} have widths of 384, 768, 1024, and 1280 and depths of 12, 12, 24, and
32. All share frozen encoders with 592M parameters, and during fine-tuning they train 39M, 106M,
326M, and 655M parameters, including the 4.6M-parameter retargeter. Table~\ref{tab:dexjoco_multitask} uses the largest
configuration, Volt-H.

\subsection{DexJoCo Evaluation Protocol}
\label{app:eval_protocol}

\textbf{Inference timing.} DexJoCo's evaluator runs asynchronously. It requests a new chunk once
fewer than 24 of the 30 actions remain and holds the robot's pose whenever the buffer empties, so
inference latency affects the executed behavior. On one B200 GPU, the Volt-L12 reference of
Table~\ref{tab:design_ablation} takes 55\,ms per forward pass, 37\,ms of it in the frozen encoder. Each simulator control step spans 20\,ms, so a
30-step chunk covers 0.6\,s. Human clips carry the same 31 frames over 1\,s
(Appendix~\ref{app:human_preprocessing}), so one predicted step spans about 33\,ms in pre-training and 20\,ms in
fine-tuning.

\textbf{Checkpoint selection.} For the Volt-H multi-task model, we consider the 20k, 40k, and 60k
fine-tuning snapshots and select on the first seed's ten-task average, which selects 60k; all component
ablations use 60k. Single-task policies are trained with the view augmentation of
Appendix~\ref{app:dexjoco} for 20k iterations (30k for bimanual hanoi and microwave) and evaluated at
that final snapshot.

\subsection{RoboDojo-Precision Evaluation Protocol}
\label{app:robodojo}

We use the official RoboDojo evaluator~\citep{robodojo2026}. Unlike the DexJoCo evaluator, it runs
synchronously. The simulator waits for each query, executes the whole action chunk, and then
queries again, so inference latency does not affect the outcome. The evaluation seed selects a fixed
set of initial layouts, and we run 50 episodes per task on seed 0, taking the layouts in order. An
episode ends at success or at a task-specific step limit, from 300 steps for insert key and deposit
coin to 1,900 steps for fasten screws, that is, 12 to 76\,s at the benchmark's 25\,Hz.

We report the success rate and the benchmark's partial-progress score, each averaged over the eight
tasks with equal weight. \methodname is fine-tuned once from the 60k pre-training checkpoint, and we
evaluate its 60k fine-tuning snapshot without further selection. The published benchmark instead
trains three seeds per policy and evaluates each for 50 trials; we use one training seed and one
evaluation seed for every method in Table~\ref{tab:robodojo_precision}. We leave out the public leaderboard
entries, since they are trained on all RoboDojo tasks rather than the Precision demonstrations
alone.

\subsection{Real-world Experiments Details}
\label{app:real}

\textbf{Banana PnP.} 
This task evaluates basic single-arm pick-and-place and, more importantly, the ability to rely on geometry when color cues are ambiguous.
We collect 54 demonstrations in which the robot picks up a banana from various positions on the table and places it in a basket at a fixed location.
We run 24 evaluation trials per method with positions sampled from those in the demonstrations, 12 with the banana and 12 with a box unseen in the demonstrations whose color closely matches the tabletop.
A trial succeeds if the object lies entirely inside the basket at the end of the episode.

\textbf{Square-peg insertion.}
This task evaluates precise grasping and tight bimanual coordination: the robot must lift a square peg and a blue hole, align them, and insert the peg into the hole. Because the peg is
slippery, an imprecise grasp allows it to tilt between the fingers, which causes the
insertion to fail. We collect 64 demonstrations with the peg and hole placed at various
positions, and run 24 evaluation trials per method with positions sampled from those in the
demonstrations. Since one end of the hole is capped, a trial is successful only if the peg
remains in the hole at the end of the episode. To capture partial progress, we additionally
score each trial on a scale from 0 to 3:
\begin{itemize}
    \item \textbf{0}: fails to lift the peg.
    \item \textbf{1}: lifts the peg but fails to lift the hole.
    \item \textbf{2}: lifts both objects but fails to insert the peg.
    \item \textbf{3}: inserts the peg, which remains in the hole until the end of the episode (success).
\end{itemize}

\subsection{Baselines}
\label{app:baselines}

\subsubsection{GR00T~N1.6}
Following the official recipe,\footnote{\url{https://github.com/NVIDIA/Isaac-GR00T}} we fine-tune
from the released GR00T~N1.6-3B checkpoint~\citep{groot_n16_2025}. We use the AdamW optimizer with a
learning rate of $1 \times 10^{-4}$, an action horizon of 30, and input images at $224 \times 224$,
with all remaining hyperparameters left at their defaults. With a global batch size of 256, we train
a separate model per task for 30K steps (single-arm) or 60K steps (bimanual) in the DexJoCo
single-task setting, and a single model across tasks for 60K steps in each of the DexJoCo multi-task
and RoboDojo settings. For evaluation, we follow the default protocols of
DexJoCo and RoboDojo-Precision.
For the EgoDex~\citep{egodex2026} pre-training experiment in Fig.~\ref{fig:pretraining_gain}, we follow
H-RDT~\citep{HRDT26} and convert the ARKit hand annotations of EgoDex
into a 44-D bimanual state per frame, consisting of the 3D position and quaternion orientation of each wrist,
which serves as the end-effector pose, and the 3D positions of the five fingertips of each hand.
Since human demonstrations carry no commanded actions, we define the action at step $t$ as the
next state, so that the model directly regresses future hand states. We pre-train for 30K steps with a global
batch size of 192, the same budget as for the other baselines.

\subsubsection{Fast-WAM}
We use the official codebase\footnote{\url{https://github.com/yuantianyuan01/FastWAM}} with the pre-trained Wan2.2-5B model. We train for 100K steps with a cosine learning-rate schedule initialized at $1\times10^{-4}$, an action horizon of 32, and a global batch size of 64. For DexJoCo, we concatenate three $224\times224$ camera views, while for RoboDojo-Precision, we concatenate three $192\times256$ views. For evaluation, we follow the default protocols of DexJoCo and RoboDojo-Precision.
In Fig.~\ref{fig:pretraining_gain},
we pretrain Fast-WAM on EgoDex with the same hand-state actions and budget as for GR00T~N1.6, with single-view $224\times224$ videos as visual supervision.

\subsubsection{PointACT}
Following the official implementation\footnote{\url{https://github.com/cshizhe/robot-PointAct}}, we initialize the VLM backbone from Qwen2.5-VL-3B-Instruct~\citep{qwen25vl2025} and keep it frozen, initialize the point encoder from the pretrained Concerto~\citep{zhang2025concerto} PTv3-Large~\citep{wu2024ptv3} checkpoint, and train the remaining modules from scratch.
We use the AdamW optimizer with a learning rate of $5\times10^{-5}$ and a cosine learning rate schedule, with random rotations of the point cloud about the gravity axis and image augmentations as in the official implementation.
With a global batch size of 256, we train a separate model per task for 30K steps in the DexJoCo single-task setting, and a single model across tasks for 60K steps in each of the DexJoCo multi-task and RoboDojo settings.
Depth maps for both benchmarks are obtained in the same manner as for PointWAM.
In Fig.~\ref{fig:pretraining_gain},
we pretrain PointACT on EgoDex with the same hand-state actions and budget as for GR00T~N1.6.

\section{Additional Quantitative Results}
\label{app:dexjoco}

\subsection{Single-Task DexJoCo Results}

\begin{table}[t]
\centering
\small
\solidrules
\setlength{\tabcolsep}{2pt}
\caption{\textbf{Single-task success rate (\%) on DexJoCo.} One policy per task in the
object-placement setting; notation as in Table~\ref{tab:dexjoco_multitask}. Our Avg. is computed per
seed over the ten retained tasks and averaged over the three seeds. ACT~\citep{act2023},
DP~\citep{diffusionpolicy2023}, $\pi_{0.5}$, and GR00T~N1.5~\citep{groot_n1_2025} results are from
\citet{dexjoco2026}; GR00T~N1.6 follows the same protocol.}
\label{tab:dexjoco_main}
\begin{tabular}{@{}lrrrrrrO}
\toprule
Task & DP-T & DP-C & ACT & $\pi_{0.5}$ & GR00T N1.5 & GR00T N1.6 & \methodname \\
\midrule
hammer nail &81.3\sd{3.1} & 58.7\sd{4.2} & 50.0\sd{7.2} & \textbf{84.7}\sd{5.0} & 67.3\sd{4.2} & 58.0\sd{2.0} & 82.0\sd{4.0} \\ %
click mouse &62.0\sd{2.0} & 74.0\sd{5.3} & 61.3\sd{3.1} & 64.7\sd{8.1} & 85.3\sd{3.1} & 66.0\sd{4.0} & \textbf{89.3}\sd{4.2} \\ %
pick bucket &83.3\sd{3.1} & 70.0\sd{2.0} & 64.0\sd{4.0} & 84.0\sd{7.2} & 72.0\sd{6.0} & 94.0\sd{2.0} & \textbf{96.7}\sd{1.2} \\ %
pinch tongs &22.7\sd{5.8} & \textbf{57.3}\sd{6.4} & 31.3\sd{3.1} & 24.0\sd{6.9} & 12.7\sd{2.3} & 33.3\sd{6.4} & 50.0\sd{12.2} \\ %
fold glasses &53.3\sd{3.1} & 54.0\sd{15.9} & 47.3\sd{11.0} & \textbf{72.0}\sd{3.5} & 27.3\sd{2.3} & 51.3\sd{8.3} & 62.0\sd{4.0} \\ %
water plant &84.0\sd{3.5} & 63.3\sd{3.1} & 47.3\sd{4.6} & 88.7\sd{3.1} & 72.7\sd{1.2} & 77.3\sd{9.0} & \textbf{94.7}\sd{1.2} \\ %
hanoi /B &24.7\sd{4.6} & 12.7\sd{3.1} & 6.0\sd{2.0} & 15.3\sd{3.1} & 0.7\sd{1.2} & 24.7\sd{1.2} & \textbf{36.0}\sd{13.9} \\ %
assembly /B &4.7\sd{3.1} & 3.3\sd{1.2} & 0.0\sd{0.0} & 5.3\sd{1.2} & 0.7\sd{1.2} & 4.0\sd{2.0} & \textbf{11.3}\sd{7.0} \\ %
microwave /B &73.3\sd{11.6} & 54.0\sd{12.5} & 66.0\sd{2.0} & 70.0\sd{3.5} & 50.7\sd{4.6} & \textbf{76.7}\sd{8.1} & 72.0\sd{10.0} \\ %
photograph /B &\textbf{56.7}\sd{4.6} & 24.0\sd{8.7} & 7.3\sd{1.2} & \textbf{56.7}\sd{5.0} & 40.7\sd{7.0} & 30.0\sd{6.9} & 24.0\sd{2.0} \\ %
\midrule
avg. &54.6\sd{1.6} & 47.1\sd{2.4} & 38.1\sd{1.5} & 56.5\sd{1.6} & 43.0\sd{1.2} & 51.5\sd{1.8} & \textbf{61.8}\sd{1.1} \\ %
\bottomrule
\end{tabular}
\end{table}

Table~\ref{tab:dexjoco_main} trains one policy per task in the object-placement setting.
\methodname averages 61.8\%, 5.3 percentage points above the strongest published baseline,
$\pi_{0.5}$ at 56.5\%, and is best on five of the ten tasks. It falls below $\pi_{0.5}$ on hammer
nail and fold glasses, below DP-C on pinch tongs, below GR00T~N1.6 and DP-T on bimanual microwave,
and below DP-T, $\pi_{0.5}$, and both GR00T models on bimanual photograph, where it reaches
only 24.0\%.

Table~\ref{tab:dexjoco_full} evaluates the same policies under full randomization, which also
varies the table height, camera pose, lighting, and texture. The single-task \methodname policies of
both tables are trained with view augmentation in 3D. Under full randomization,
\methodname averages 62.9\% and is best on all ten tasks. Every baseline loses 9.4 to 32.8 percentage points, so the margin of \methodname over the strongest
baseline widens from 5.3 to 22.2 percentage points.

\begin{table}[t]
\centering
\small
\solidrules
\setlength{\tabcolsep}{2pt}
\caption{\textbf{Single-task success rate (\%) on DexJoCo under full randomization.}
Object placement, table height, camera pose, lighting, and texture vary.
ACT, DP, $\pi_{0.5}$, and GR00T~N1.5 results are from \citet{dexjoco2026};
notation follows Table~\ref{tab:dexjoco_main}. \methodname evaluates the same viewpoint-augmented
policies as Table~\ref{tab:dexjoco_main}, trained as described in Appendix~\ref{app:dexjoco}.}
\label{tab:dexjoco_full}
\begin{tabular}{@{}lrrrrrrO}
\toprule
Task & DP-T & DP-C & ACT & $\pi_{0.5}$ & GR00T N1.5 & GR00T N1.6 & \methodname \\
\midrule
hammer nail &18.7\sd{1.2} & 19.3\sd{3.1} & 22.7\sd{6.1} & 17.3\sd{5.0} & 38.7\sd{8.3} & 26.7\sd{7.0} & \textbf{77.3}\sd{6.4} \\
click mouse &25.3\sd{8.1} & 34.7\sd{4.2} & 48.7\sd{5.0} & 54.7\sd{7.0} & 74.0\sd{2.0} & 50.7\sd{4.2} & \textbf{94.7}\sd{1.2} \\
pick bucket &58.7\sd{15.0} & 68.0\sd{3.5} & 36.0\sd{5.3} & 78.7\sd{6.1} & 69.3\sd{6.1} & 85.3\sd{3.1} & \textbf{96.0}\sd{4.0} \\
pinch tongs &18.7\sd{3.1} & 28.7\sd{11.7} & 23.3\sd{7.0} & 18.7\sd{1.2} & 5.3\sd{2.3} & 38.0\sd{2.0} & \textbf{47.3}\sd{5.8} \\
fold glasses &11.3\sd{1.2} & 15.3\sd{7.6} & 7.3\sd{3.1} & 39.3\sd{3.1} & 20.7\sd{3.1} & 38.0\sd{2.0} & \textbf{66.0}\sd{10.6} \\
water plant &56.0\sd{8.7} & 54.0\sd{5.3} & 52.7\sd{8.1} & 75.3\sd{6.4} & 66.0\sd{5.3} & 76.7\sd{4.2}  & \textbf{94.0}\sd{2.0} \\
hanoi /B &0.7\sd{1.2} & 9.3\sd{6.1} & 4.7\sd{2.3} & 15.3\sd{2.3} & 0.0\sd{0.0} & 22.0\sd{6.0} & \textbf{47.3}\sd{6.1} \\
assembly /B &0.0\sd{0.0} & 0.0\sd{0.0} & 0.0\sd{0.0} & 0.0\sd{0.0} & 1.3\sd{1.2} & 0.0\sd{0.0} & \textbf{3.3}\sd{1.2} \\
microwave /B &21.3\sd{4.6} & 62.7\sd{6.4} & 50.0\sd{6.9} & 54.7\sd{6.1} & 42.0\sd{7.2} & 52.7\sd{7.0} & \textbf{74.0}\sd{4.0} \\
photograph /B &7.3\sd{1.2} & 8.7\sd{4.2} & 3.3\sd{1.2} & 21.3\sd{2.3} & 18.7\sd{7.0} & 16.7\sd{8.1} & \textbf{29.3}\sd{1.2} \\
\midrule
avg. &21.8 & 30.1 & 24.9 & 37.5 & 33.6 & 40.7\sd{2.6} & \textbf{62.9}\sd{2.2} \\ %
\bottomrule
\end{tabular}
\end{table}

\begin{table}[t]
\centering
\caption{\textbf{Additional component ablations.} Same Volt-L12 reference and
protocol as Table~\ref{tab:design_ablation}; shading marks the reference setting, and $M$ is the average number of vision tokens per sample.}
\label{tab:design_ablation_appendix}

\abcell[lrc][3]{a}{patch size}{$K_v$ & $M$ & SR (\%)}{%
3 & 1150 & 67.1\sd{1.6} \\
\dfltrowiii{5}{452}{64.4\sd{1.7}}
7 & 249 & 44.5\sd{1.6} \\}\hspace{0.3in}%
\abcell{b}{encoder fine-tuning}{case & SR (\%)}{%
\dfltrow{frozen}{64.4\sd{1.7}}
fine-tuned & 54.9\sd{0.9} \\}
\end{table}

\FloatBarrier

\subsection{Patch Size and Encoder Fine-Tuning}
\label{app:additional_ablations}

Table~\ref{tab:design_ablation_appendix} varies the patch size and encoder fine-tuning from the
Volt-L12 reference of Table~\ref{tab:design_ablation}. With 1\,cm voxels, reducing $K_v$ from 5 to 3
raises SR from 64.4\% to 67.1\% but costs 1.63$\times$ the training step time and 1.66$\times$ the
peak memory, while $K_v=7$ lowers SR to 44.5\% (a). We therefore keep $K_v=5$, which balances
manipulation success against training time and memory. Fine-tuning the Mosaic3D encoder instead of
keeping it frozen lowers SR from 64.4\% to 54.9\% (b), so the encoder stays frozen in all other
experiments of this paper.

\FloatBarrier
\section{Additional Qualitative Results}
\label{app:qualitative}

\begin{figure}[p]
\centering
\newlength{\eqw}\setlength{\eqw}{\dimexpr(\linewidth-20pt)/4\relax}
\newlength{\eqh}\setlength{\eqh}{0.6\eqw}
\newcommand{\eimg}[1]{\includegraphics[width=\eqw,height=\eqh]{figures/#1}}
\newcommand{\erow}[2]{\makebox[\linewidth][c]{\makebox[12pt][c]{\rotatebox{90}{\makebox[\eqh][c]{\footnotesize #1}}}\hspace{2pt}%
  \eimg{qual_app/success/#2_t0.jpg}\hspace{2pt}\eimg{qual_app/success/#2_t1.jpg}\hspace{2pt}\eimg{qual_app/success/#2_t2.jpg}\hspace{2pt}\eimg{qual_app/success/#2_t3.jpg}}\\[2pt]}
\newcommand{\ehdr}[1]{\makebox[\eqw][c]{\footnotesize #1}}
\makebox[\linewidth][c]{\makebox[14pt]{}\ehdr{$t=0$}\hspace{2pt}\ehdr{$t=t_\mathrm{end}/3$}\hspace{2pt}%
  \ehdr{$t=2t_\mathrm{end}/3$}\hspace{2pt}\ehdr{$t=t_\mathrm{end}$}}\\[1pt]
\erow{hammer nail}{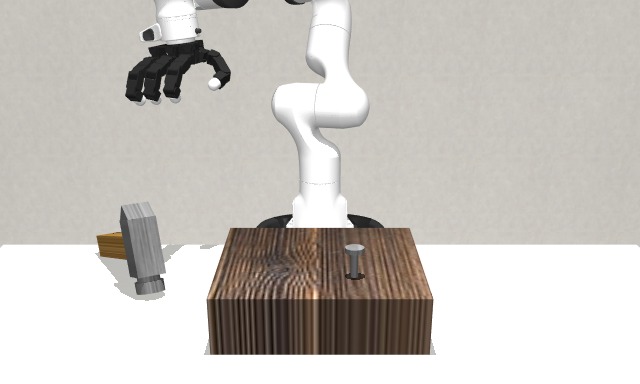}
\erow{click mouse}{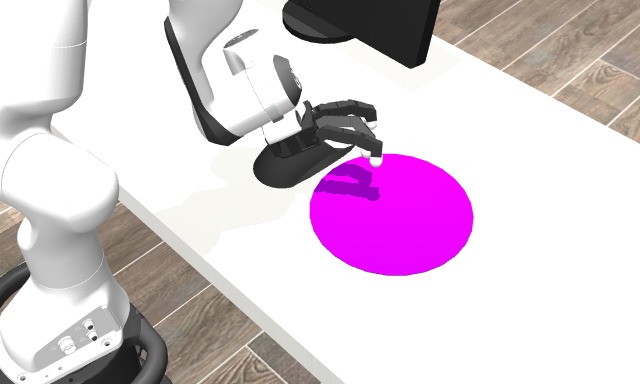}
\erow{pick bucket}{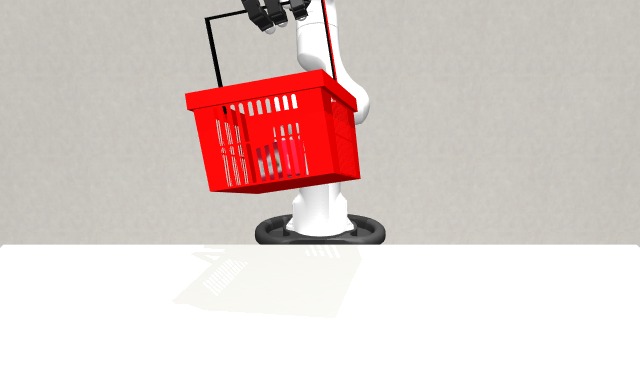}
\erow{pinch tongs}{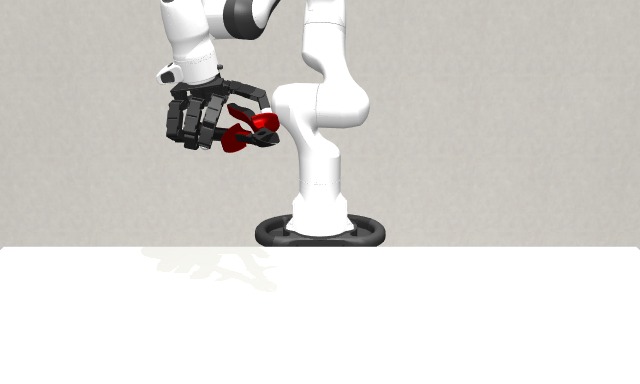}
\erow{fold glasses}{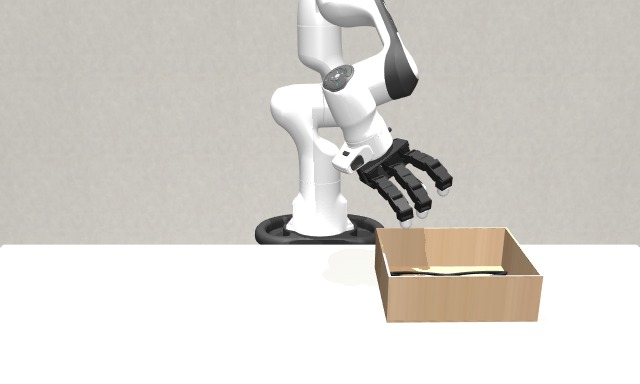}
\erow{water plant}{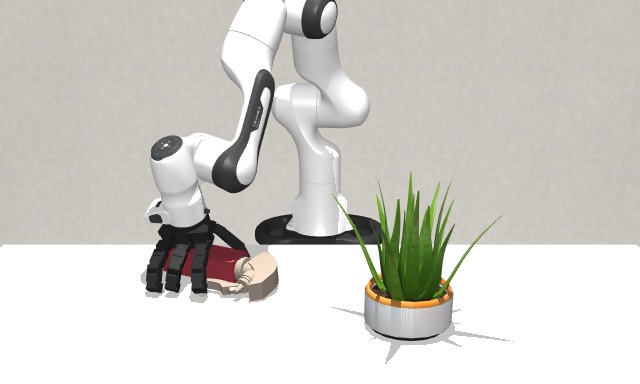}
\erow{hanoi /B}{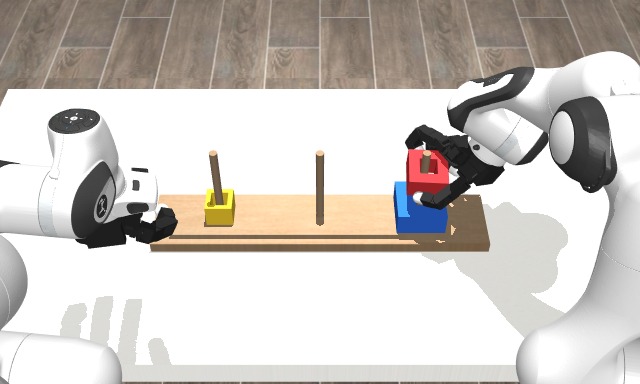}
\erow{assembly /B}{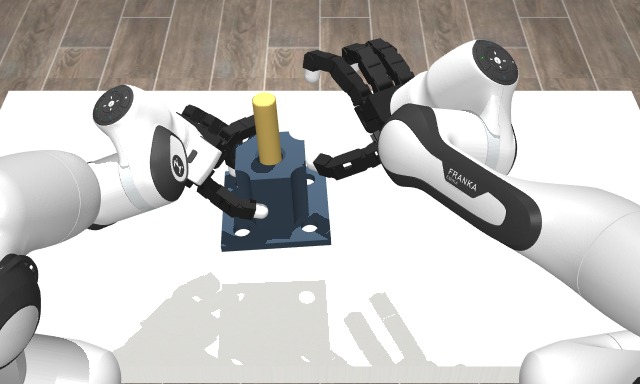}
\erow{microwave /B}{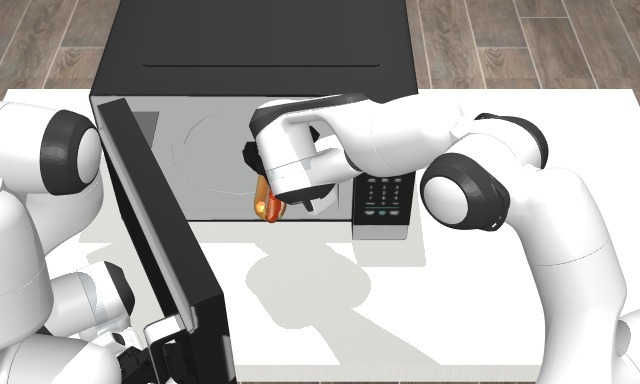}
\erow{photograph /B}{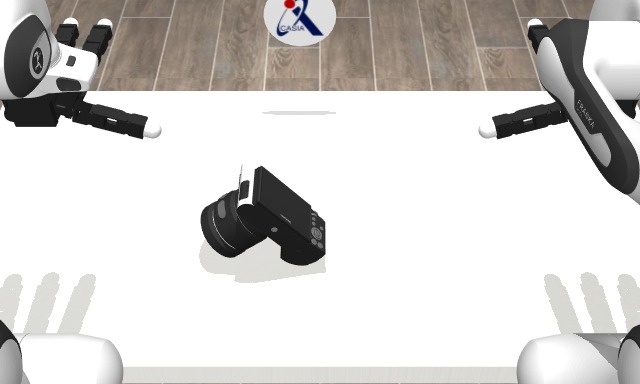}
\caption{\textbf{Successful \methodname rollouts on DexJoCo.} One rollout per task from the
multi-task policy, the first success among evaluation seeds 0--7, shown at four points up to its own
$t_\mathrm{end}$.}
\label{fig:app_qual_dexjoco}
\end{figure}

\begin{figure}[p]
\centering
\newlength{\fqw}\setlength{\fqw}{\dimexpr(\linewidth-20pt)/4\relax}
\newlength{\fqh}\setlength{\fqh}{0.6\fqw}
\newcommand{\fimg}[1]{\includegraphics[width=\fqw,height=\fqh]{figures/#1}}
\newcommand{\frow}[2]{\makebox[\linewidth][c]{\makebox[12pt][c]{\rotatebox{90}{\makebox[\fqh][c]{\footnotesize #1}}}\hspace{2pt}%
  \fimg{qual_app/fail/#2_t0.jpg}\hspace{2pt}\fimg{qual_app/fail/#2_t1.jpg}\hspace{2pt}\fimg{qual_app/fail/#2_t2.jpg}\hspace{2pt}\fimg{qual_app/fail/#2_t3.jpg}}\\[2pt]}
\newcommand{\fhdr}[1]{\makebox[\fqw][c]{\footnotesize #1}}
\makebox[\linewidth][c]{\makebox[14pt]{}\fhdr{$t=0$}\hspace{2pt}\fhdr{$t=t_\mathrm{end}/3$}\hspace{2pt}%
  \fhdr{$t=2t_\mathrm{end}/3$}\hspace{2pt}\fhdr{$t=t_\mathrm{end}$}}\\[1pt]
\frow{hammer nail}{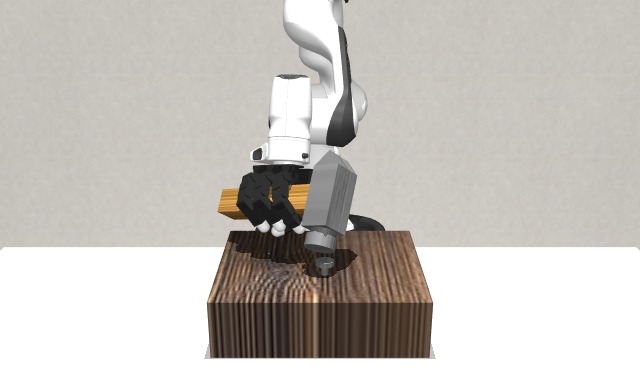}
\frow{click mouse}{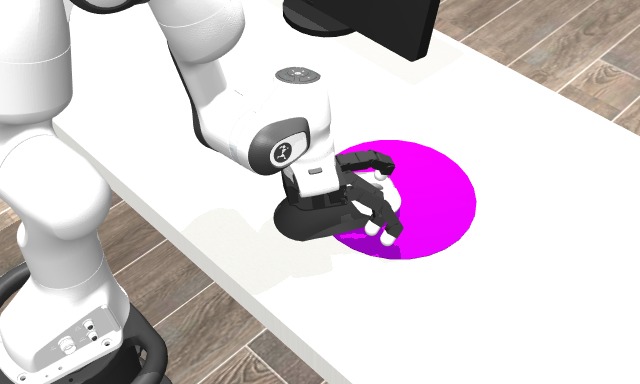}
\frow{pick bucket}{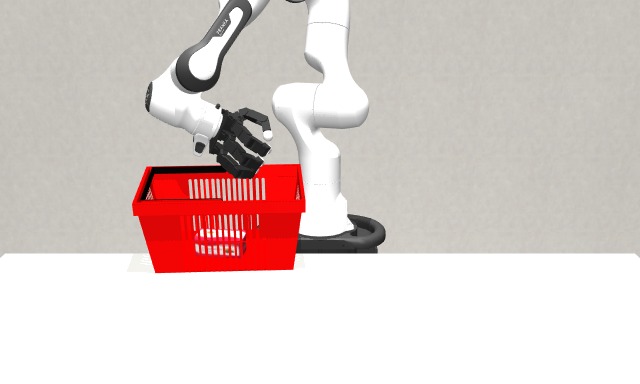}
\frow{pinch tongs}{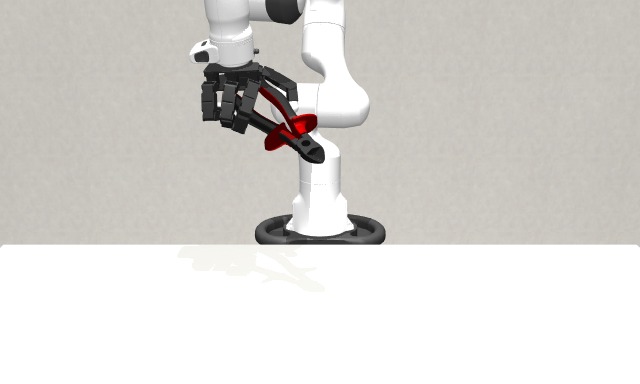}
\frow{fold glasses}{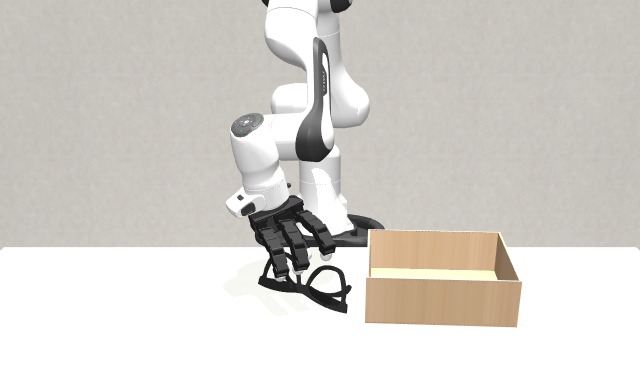}
\frow{hanoi /B}{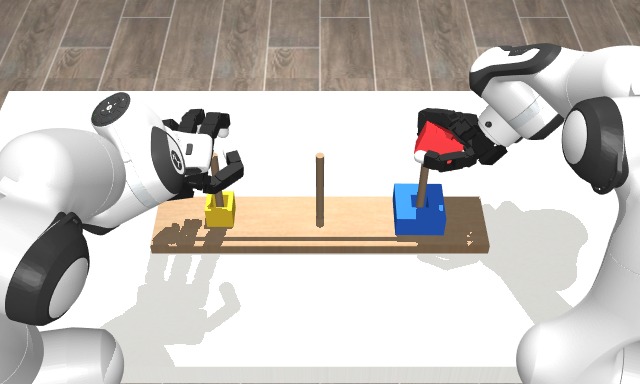}
\frow{assembly /B}{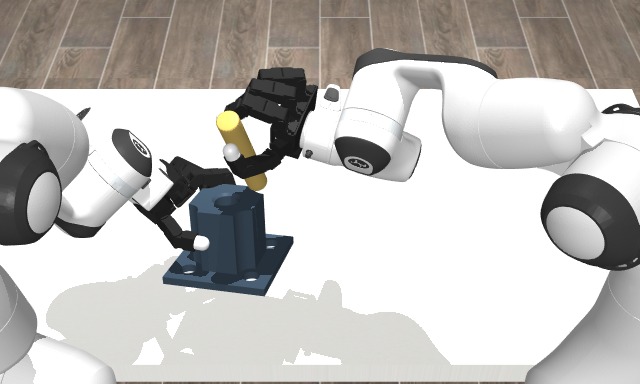}
\frow{microwave /B}{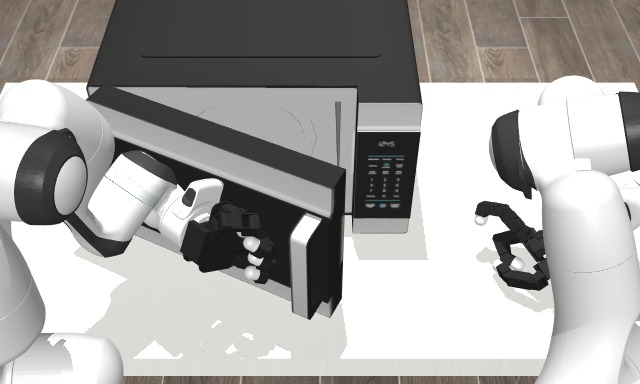}
\frow{photograph /B}{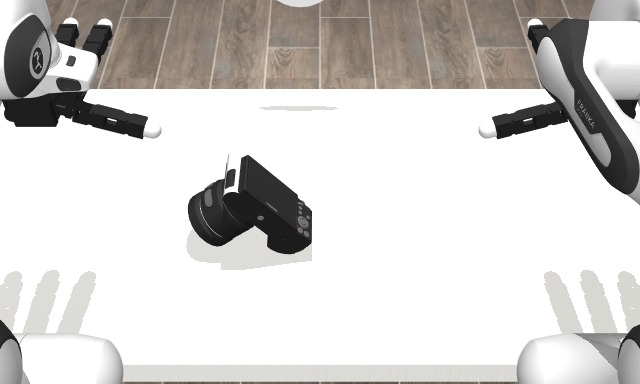}
\caption{\textbf{Failed \methodname rollouts on DexJoCo.} One failed rollout per task from the
multi-task policy, the first failure among the same rollouts as Figure~\ref{fig:app_qual_dexjoco},
shown at four points up to its own $t_\mathrm{end}$. Water plant has no row: all 16 of its rollouts
(seeds 0--15) succeed.}
\label{fig:app_qual_dexjoco_fail}
\end{figure}

\begin{figure}[p]
\centering
\definecolor{rolloutsuccess}{HTML}{2CA02C}
\definecolor{rolloutfailure}{HTML}{D62728}
\newlength{\dqw}\setlength{\dqw}{\dimexpr(\linewidth-20pt)/4\relax}
\newlength{\dqh}\setlength{\dqh}{0.6\dqw}
\newcommand{\dimg}[1]{\includegraphics[width=\dqw,height=\dqh]{figures/qual_app/robodojo/#1}}
\newcommand{\dlast}[2]{\begin{tikzpicture}[baseline=(img.base)]
  \node[inner sep=0pt] (img) {\dimg{#2}};
  \draw[#1, line width=1.2pt] ([shift={(0.6pt,0.6pt)}]img.south west) rectangle ([shift={(-0.6pt,-0.6pt)}]img.north east);
\end{tikzpicture}}
\newcommand{\drow}[3]{\makebox[\linewidth][c]{\makebox[12pt][c]{\rotatebox{90}{\makebox[\dqh][c]{\footnotesize #1}}}\hspace{2pt}%
  \dimg{#3_t0.jpg}\hspace{2pt}\dimg{#3_t1.jpg}\hspace{2pt}\dimg{#3_t2.jpg}\hspace{2pt}\dlast{#2}{#3_t3.jpg}}\\[2pt]}
\newcommand{\dhdr}[1]{\makebox[\dqw][c]{\footnotesize #1}}
\makebox[\linewidth][c]{\makebox[14pt]{}\dhdr{$t=0$}\hspace{2pt}\dhdr{$t=t_\mathrm{end}/3$}\hspace{2pt}%
  \dhdr{$t=2t_\mathrm{end}/3$}\hspace{2pt}\dhdr{$t=t_\mathrm{end}$}}\\[1pt]
\drow{fasten screws}{rolloutfailure}{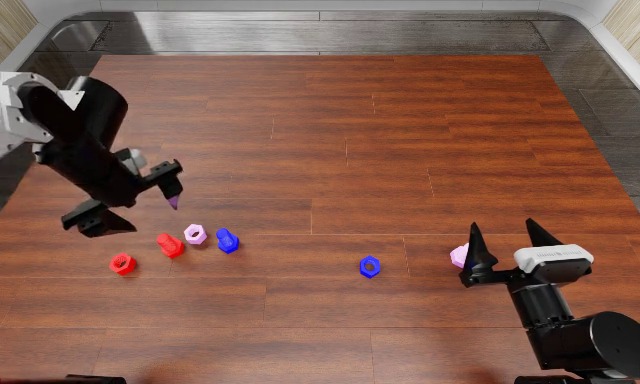}
\drow{plug in charger}{rolloutfailure}{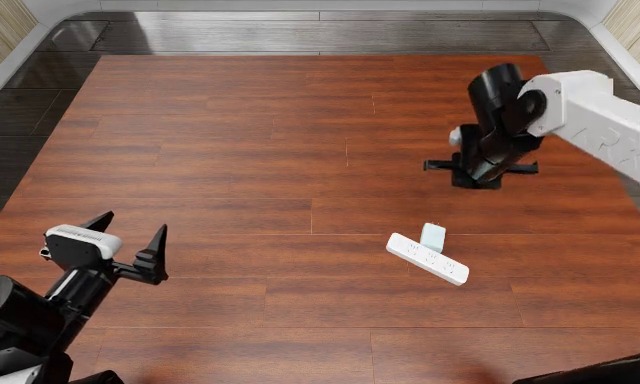}
\drow{insert tubes}{rolloutsuccess}{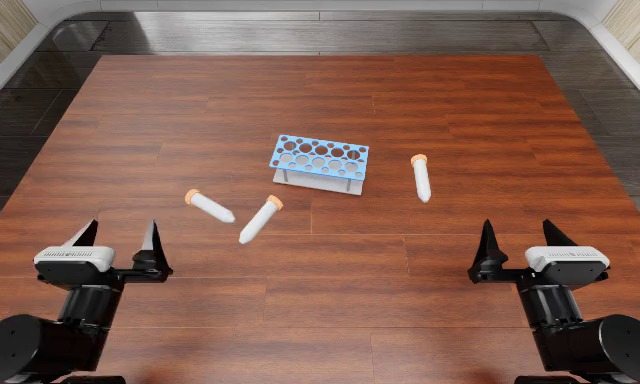}
\drow{pour balls}{rolloutsuccess}{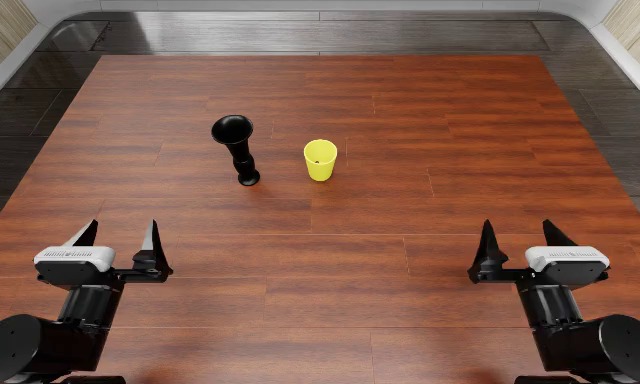}
\drow{play xylophone}{rolloutsuccess}{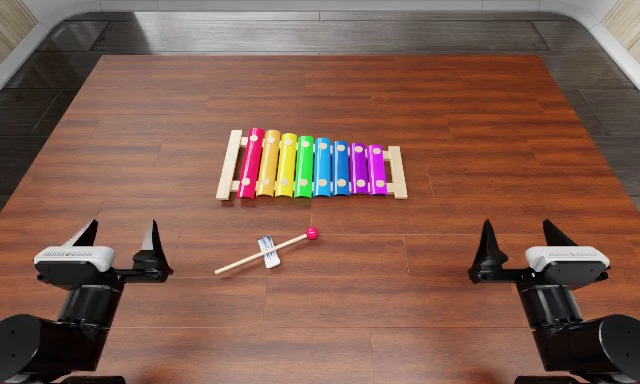}
\drow{deposit coin}{rolloutsuccess}{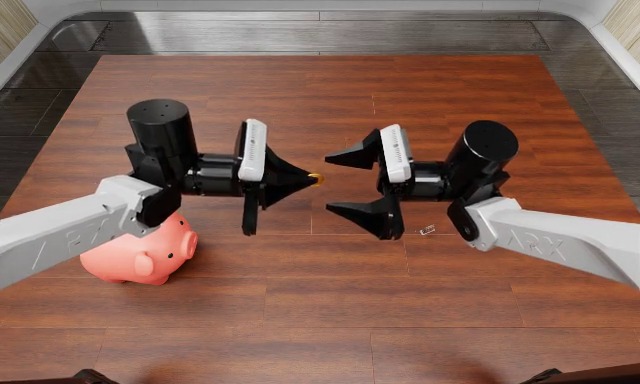}
\drow{insert key}{rolloutfailure}{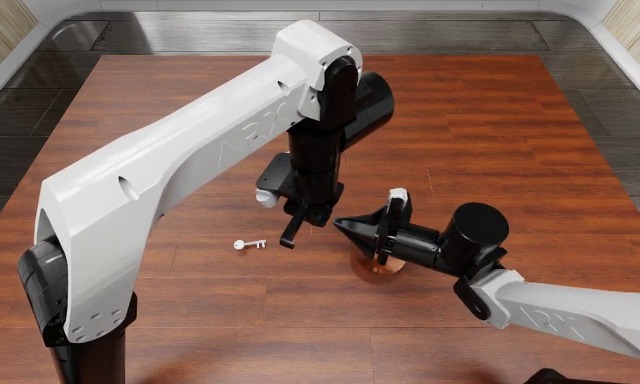}
\drow{build tower}{rolloutsuccess}{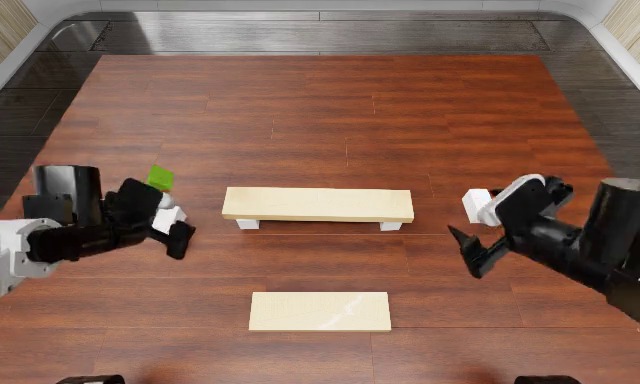}
\caption{\textbf{\methodname rollouts on RoboDojo-Precision.} One rollout per task, shown at four
points up to its own $t_\mathrm{end}$: a successful one (\textcolor{rolloutsuccess}{green} border) for
the five tasks \methodname solves at least once, and a failed one (\textcolor{rolloutfailure}{red}
border) for the other three tasks.}
\label{fig:app_qual_robodojo}
\end{figure}

\begin{figure}[p]
\centering
\newlength{\rqw}\setlength{\rqw}{\dimexpr(\linewidth-42pt)/4\relax}
\newlength{\rqh}\setlength{\rqh}{0.6\rqw}
\definecolor{rolloutsuccess}{HTML}{2CA02C}
\definecolor{rolloutfailure}{HTML}{D62728}
\newcommand{\rimg}{\includegraphics[width=\rqw,height=\rqh]{example-image-duck}}
\newcommand{\rfimg}[1]{\includegraphics[width=\rqw,height=\rqh]{figures/qual_real/#1}}
\newcommand{\rlast}[2]{\begin{tikzpicture}[baseline=(img.base)]
  \node[inner sep=0pt] (img) {\rfimg{#2}};
  \draw[#1, line width=1.2pt] ([shift={(0.6pt,0.6pt)}]img.south west) rectangle ([shift={(-0.6pt,-0.6pt)}]img.north east);
\end{tikzpicture}}
\newcommand{\rlabel}[1]{\makebox[12pt][c]{\rotatebox{90}{\makebox[\rqh][c]{\footnotesize #1}}}\hspace{2pt}}
\newcommand{\rrow}[1]{\rlabel{#1}\rimg\hspace{2pt}\rimg\hspace{2pt}\rimg\hspace{2pt}\rimg}
\newcommand{\rrec}[3]{\rlabel{#1}\rfimg{#3_t0.png}\hspace{2pt}\rfimg{#3_t1.png}\hspace{2pt}%
  \rfimg{#3_t2.png}\hspace{2pt}\rlast{#2}{#3_t3.png}}
\newcommand{\rtask}[2]{\makebox[\linewidth][c]{%
  \makebox[20pt][c]{\rotatebox{90}{\makebox[\dimexpr3\rqh+4pt][c]{\footnotesize\begin{tabular}{@{}c@{}}#1\end{tabular}}}}\hspace{2pt}%
  \begin{tabular}[b]{@{}l@{}}#2\end{tabular}}\\[4pt]}
\newcommand{\rhdr}[1]{\makebox[\rqw][c]{\footnotesize #1}}
\makebox[\linewidth][c]{\makebox[36pt]{}\rhdr{$t=0$}\hspace{2pt}\rhdr{$t=t_\mathrm{end}/3$}\hspace{2pt}%
  \rhdr{$t=2t_\mathrm{end}/3$}\hspace{2pt}\rhdr{$t=t_\mathrm{end}$}}\\[1pt]
\rtask{banana pick-and-place\\banana (in-domain)}{\rrec{GR00T~N1.6}{rolloutsuccess}{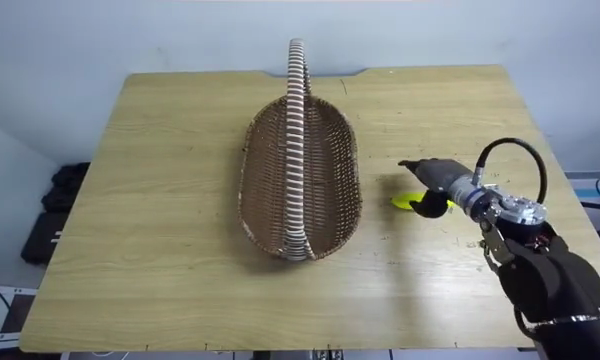}\\[-0.5pt]%
  \rrec{$\pi_{0.5}$}{rolloutsuccess}{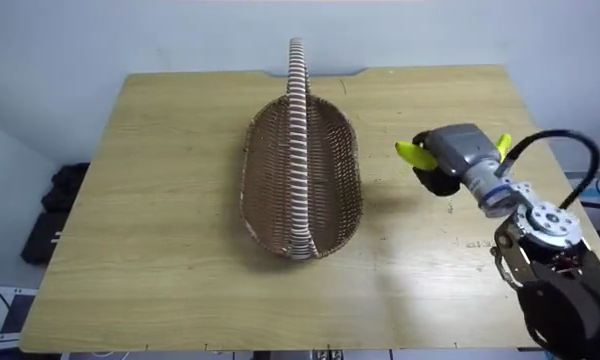}\\[-0.5pt]\rrec{\methodname}{rolloutsuccess}{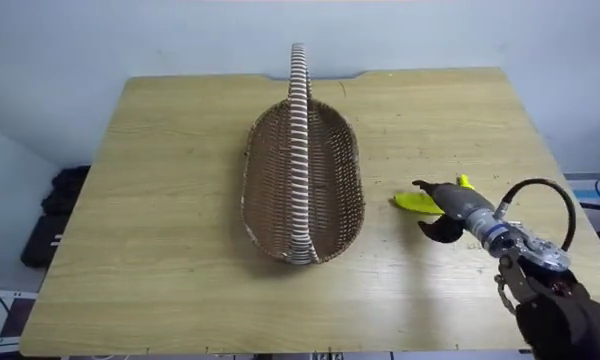}}
\rtask{banana pick-and-place\\unseen box (out-of-domain)}{\rrec{GR00T~N1.6}{rolloutfailure}{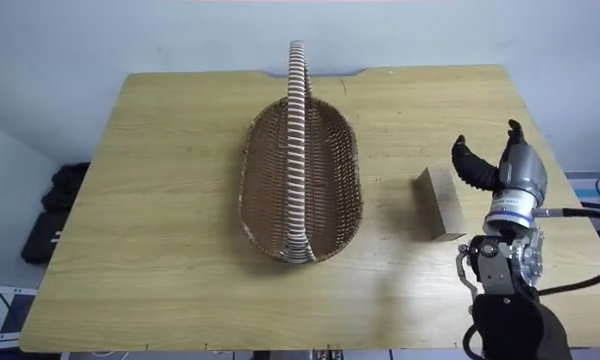}\\[-0.5pt]%
  \rrec{$\pi_{0.5}$}{rolloutfailure}{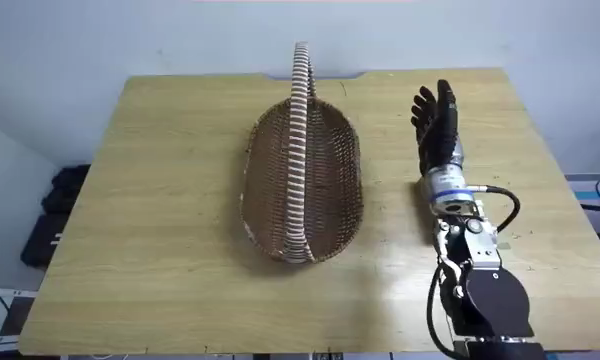}\\[-0.5pt]\rrec{\methodname}{rolloutsuccess}{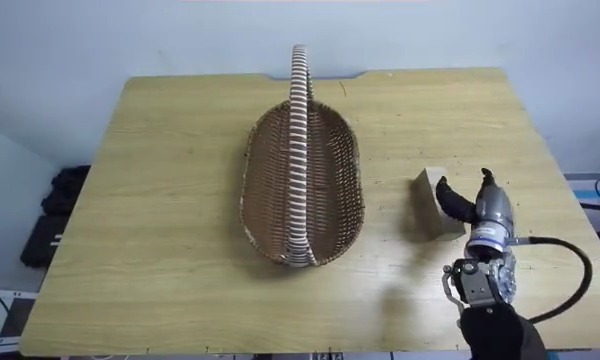}}
\rtask{square-peg insertion\\\mbox{}}{\rrec{GR00T~N1.6}{rolloutfailure}{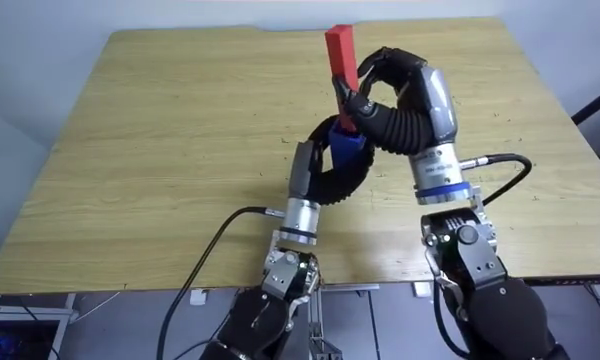}\\[-0.5pt]%
  \rrec{$\pi_{0.5}$}{rolloutfailure}{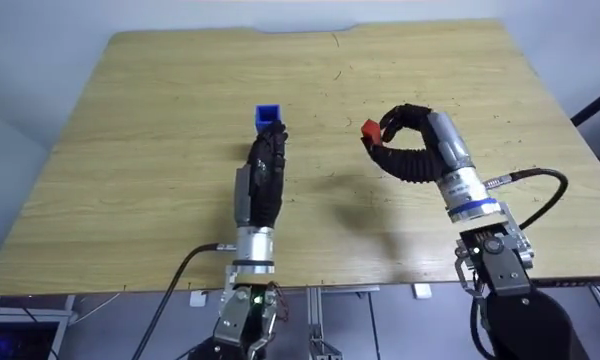}\\[-0.5pt]\rrec{\methodname}{rolloutsuccess}{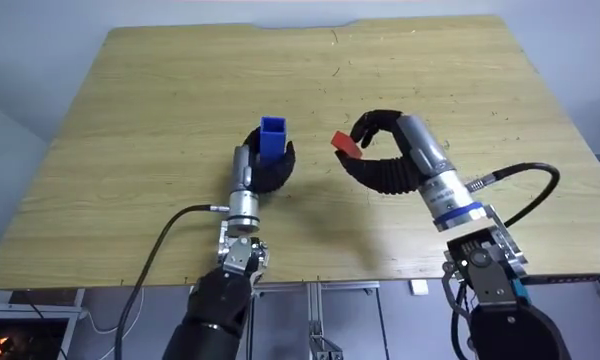}}
\caption{\textbf{Real-robot rollouts.} GR00T~N1.6, $\pi_{0.5}$, and \methodname on the two tasks,
each shown at four points up to its own $t_\mathrm{end}$ with a \textcolor{rolloutsuccess}{green} border for success
and a \textcolor{rolloutfailure}{red} border for failure. Banana pick-and-place shows one trial
each with the in-domain banana and the out-of-domain unseen box.}
\label{fig:app_qual_real}
\end{figure}

Figures~\ref{fig:app_qual_dexjoco} and~\ref{fig:app_qual_dexjoco_fail} show successful and failed
\methodname rollouts on the ten DexJoCo tasks. For each task, we take the first success and the
first failure among eight rollouts, one episode on each of evaluation seeds 0--7; water plant
succeeds in all sixteen rollouts we ran, so it has no failure row. In most failures, \methodname
grasps the object but stops short of the last step until the time limit. It holds the hammer above
the nail, the peg beside the hole, the glasses next to the box, and the camera without pressing the
shutter. Figure~\ref{fig:app_qual_robodojo} shows rollouts on the eight RoboDojo-Precision tasks, and
Figure~\ref{fig:app_qual_real} compares GR00T~N1.6, $\pi_{0.5}$, and \methodname on the two
real-robot tasks.

\end{document}